\documentclass[oneside,11pt]{article}

\newtheorem{theorem}{Theorem}
\newtheorem{lemma}{Lemma}
\newtheorem{proposition}{Proposition}
\newcommand{\Proof}{\NI
                    {\bf Proof.}\ }

\usepackage{url}
\usepackage{latexsym} 

\input{epsf}

\usepackage{amsmath}
\usepackage{amssymb}
\usepackage{amsfonts}
\usepackage{graphicx}
\def\0{{\bf 0}}
\def\1{{\bf 1}}

\def\smallromani{\renewcommand{\theenumi}{\roman{enumi}}
\renewcommand{\labelenumi}{(\theenumi)}}

\newcommand{\oldbfe}[1]{\begin{bfseries}\emph{#1}\end{bfseries}}

\newcommand{\ES}{\mbox{$\emptyset$}}

\newcommand{\myra}{\mbox{$\:\rightarrow\:$}}

\newcommand{\La}{\mbox{$\:\Leftarrow\:$}}
\newcommand{\Ra}{\mbox{$\:\Rightarrow\:$}}

\newcommand{\A}{\mbox{$\ \wedge\ $}}

\newcommand{\sse}{\mbox{$\:\subseteq\:$}}

\newcommand{\fa}{\mbox{$\forall$}}
\newcommand{\te}{\mbox{$\exists$}}

\newcommand{\LL}{\mbox{$\ldots$}}

\newcommand{\C}[1]{\mbox{$\{{#1}\}$}}           % curly braces

\newcommand{\NI}{\noindent}
\newcommand{\HB}{\hfill{$\Box$}}
\newcommand{\VV}{\vspace{5 mm}}

\newcommand{\II}{\vspace{2 mm}}

\newcommand{\szkew}[1]{\relax \setbox0=\hbox{\kern -24pt $\displaystyle#1$\kern 0pt }%
\box0}
{\catcode`\@=11 \global\let\ifjusthvtest@=\iffalse}
\newcounter{oldmycaption}

\usepackage{color}
\usepackage{sgame}
\usepackage{amsmath}
\usepackage{colordvi}
\usepackage{latexsym}

\begin{document}
\date{}
\title{\Large\bf Comparing the notions of optimality in CP-nets, strategic games and soft constraints}
\author{Krzysztof R. Apt$^{1,2}$, Francesca Rossi$^{3}$, and Kristen
  Brent Venable$^{3}$\\ 
$^{1}$ {\small CWI Amsterdam,} \\
{\small Kruislaan 413 1098 SJ, Amsterdam, the Netherlands}\\
E-mail: {\tt  apt@cwi.nl} \\
$^{2}${\small University of Amsterdam,}\\ 
{\small Plantage Muidergracht 24 1018 TV, Amsterdam, the Netherlands}\\ 
$^{3}${\small Department of Pure and Applied Mathematics,}\\
\small{University of Padova,}\\ 
{\small Via Trieste, 63 - 35121, Padova, Italy}\\
E-mails:{\tt \{frossi,kvenable\}@math.unipd.it}
{\small preferences,
CP nets, soft constraints}\\
{\bf AMS MOS Classification:} {\small 91B10, 91B50, 68T01, 68T30}
}

\maketitle

\begin{abstract}
 
%new short abstract
  The notion of optimality naturally arises in many areas of applied
  mathematics and computer science concerned with decision making.
  Here we consider this notion in the context of three formalisms used
  for different purposes in reasoning about multi-agent systems:
  strategic games, CP-nets, and soft constraints.  To relate the
  notions of optimality in these formalisms we introduce a natural
  qualitative modification of the notion of a strategic game.  We show
  then that the optimal outcomes of a CP-net are exactly the Nash
  equilibria of such games. This allows
  us to use the techniques of game theory to search for optimal
  outcomes of CP-nets and vice-versa, to use techniques developed for
  CP-nets to search for Nash equilibria of the considered games.
  Then, we relate the notion of optimality used in the area of soft
  constraints to that used in a generalization of strategic games,
  called graphical games.  In particular we prove that for a natural class of soft
  constraints that includes weighted constraints every optimal
  solution is both a Nash equilibrium and Pareto efficient joint strategy.
  For a natural mapping in the other direction we show that Pareto efficient joint strategies
  coincide with the optimal solutions of soft constraints.

\end{abstract}

\section{Introduction}

%\subsection{Background}

The concept of optimality is prevalent in many areas of applied
mathematics and computer science.  It is of relevance whenever we need
to choose among several alternatives that are not equally preferable.
For example, in constraint optimization, each solution of a constraint
satisfaction problem has a quality level associated with it and the
aim is to choose an optimal solution, that is, a solution with an
optimal quality level.  In turn, in strategic games, two concepts of
optimality have been commonly used: Nash equilibrium and Pareto
efficient outcome.

Some formalisms proposed in AI employ `their own' concept of an
optimal outcome.  The aim of this paper is to clarify the status of
such notions of optimality used in CP-nets and soft constraints. To
this end we use tools and techniques from game theory, more
specifically theory of strategic games.

This allows us to gain new insights into the relationship between
these formalisms which hopefully will lead to further
cross-fertilization among these three different approaches to
modelling optimality.

\subsection{Background}

\emph{Game theory}, notably the theory of \emph{strategic games}, forms
one of the main tools in the area of multi-agent systems since they
formalize in a simple and powerful way the idea that the agents
interact with each other while pursuing their own interests.  Each
agent has a set of strategies and a payoff function on the set of
joint strategies. The agents choose their strategies simultaneously
with the aim of maximizing one's payoff.  

The most commonly used
concept of optimality is that of a Nash equilibrium. Intuitively, it
is an outcome that is optimal for each player under the assumption
that only he may reconsider his action.  Another concept of optimality
is that of Pareto efficient joint strategies, which are those in which no
player can improve his payoff without decreasing the payoff of some
other player.  Sometimes it is useful to consider constrained Nash
equilibria, that is, Nash equilibria that satisfy some additional
requirements, see e.g.~\cite{constNash}.  For example, Pareto efficient Nash
equilibria are Nash equilibria which are also Pareto efficient among
the Nash equilibria.

In turn, \emph{CP-nets} (Conditional Preference nets) are an elegant
formalism for representing conditional and qualitative preferences,
see \cite{BBHP.UAI99,BBHP.journal}.  They model such preferences under
a {\em ceteris paribus} (that is, `all else being equal') assumption.
A CP-net exploits the idea of conditional independence to provide a compact 
representation of preference problems.
Preference elicitation in such a framework appears to
be natural and intuitive.

Research on CP-nets has been focused on their modeling capabilities 
and on
algorithms for solving various natural problems related to their use.
Also, computational complexity of these problems was extensively
studied.  
%An outcome of a CP-net is an assignment of values to its
%variables. 
One of the fundamental problems is that of finding an
optimal outcome, i.e., one that cannot be improved in the presence of the
adopted preference statements. This is in general a complex problem
since it was found that finding optimal outcomes and testing for their
existence is in general NP-hard, see~\cite{BBHP.UAI99,BBHP.journal}. 
In contrast, for so-called acyclic
CP-nets this is an easy problem which can be solved by a linear time
algorithm, see~\cite{BBHP.UAI99,BBHP.journal}.

Finally, \emph{soft constraints}, see e.g. \cite{jacm}, 
%brent r1
are a quantitative formalism   
which allow us to express constraints and preferences.
While constraints state which combinations of variable values are
acceptable, soft constraints
allow for several levels of acceptance.
% in the presence of constraints and uncertainty.
An example are fuzzy constraints, see \cite{fuzzy1} and
\cite{ruttkay-fuzzy}, where acceptance levels 
are between 0 and 1, and where the quality of a solution is the
minimal level over all the constraints. An optimal
solution is the one with the highest quality.
The research in this
area has dealt mainly with the algorithms for finding optimal solutions
and with the relationship between modelling formalisms,
see~\cite{MRS06}. 

\subsection{Main results}
%Contributions}

We consider the notions of optimality in
two preference modelling frameworks, that is, CP-nets and soft constraints,
and in strategic games. Although apparently there is no connection 
among these different ways of modelling preferences, we show that in fact
there is a strong relationship. This is surprising and interesting on its own.
Moreover, it might be exploited for a cross-fertilization among these
three frameworks.

In particular, we start by considering the relationship between CP-nets and strategic
games, and we show how game-theoretic techniques
can be fruitfully used to study CP-nets.
Our approach is based on the observation that the ceteris-paribus
principle, typical of CP-nets, implies that an optimal outcome is
worsened if a worsening change (to some variable) is made. This is
exactly the idea behind Nash equilibria and the desired results
easily follow once this observation is made formal by introducing an
appropriate modification of strategic games.  In this modification
each player has at his disposal a preference relation on his set of
strategies, parametrized by a joint strategy of his opponents. We call
such games \emph{strategic games with parametrized preferences}.

The cornerstone of our approach are two results closely relating
CP-nets to such games. They show that the optimal outcomes of a CP-net
are exactly the Nash equilibria of an appropriately defined strategic
game with parametrized preferences. This allows us to transfer
techniques of game theory to CP-nets, and vice-versa. 

%brent r1
%To find Nash equilibria 
In strategic games techniques
have been studied which iteratively reduce the game by eliminating some players'
strategies, thus obtaining a smaller game while maintaining 
its Nash equilibria. 
In \cite{GKZ90}, for example,  interesting results
concerning the order in which such reductions are applied are described.  
We introduce two
counterparts of such game-theoretic techniques that allow us to reduce
a CP-net while maintaining its optimal outcomes. We also introduce a
method of simplifying a CP-net by eliminating so-called redundant
variables from the variables parent sets. Both techniques simplify the
search for optimal outcomes in a CP-net.

In the other direction, we can use the techniques developed to reason
about optimal outcomes of a CP-net to search for Nash equilibria of
strategic games with parametrized preferences.  We illustrate this
point by introducing the notion of a hierarchical game with
parametrized preferences and by explaining that such games have a
unique Nash equilibrium that can be found in linear time.

In the final part of the paper we consider the relationship between
strategic games and soft constraints, such as fuzzy, 
weighted and hard constraints. The appropriate notion of a strategic game 
is here that of a \emph{graphical game}, see \cite{KLS01}.
This is due to the fact that (soft) constraints
usually involve only a small subset of the problem variables.
This is in analogy with the fact that in a graphical game a
player's payoff function depends only on a (usually small) 
number of other players.

We consider a natural mapping that associates with each soft constraint
satisfaction problem (in short, a soft CSP or an SCSP) a graphical game. This
mapping creates a direct corresponce between constraints and players'
neighbourhoods.  We show that, when using such a mapping, in general no
relation exists between the notions of optimal solutions in soft CSPs
and Nash equilibria in the corresponding games.  On the other hand,
for the class of strictly monotonic SCSPs (which includes in particular
weighted constraints), every optimal solution corresponds to both a Nash
equilibrium and Pareto efficient joint strategy.  
We also show that this mapping, when applied to a
consistent CSP (that is, a satisfiable hard constraint satisfaction
problem), defines a bijection between the solutions of the CSP and the
set of joint strategies that are both Nash equilibria and Pareto efficient.

The latter holds in general, and not just for a subclass, if we
consider a mapping from graphical games to soft CSPs which is
independent of the constraint structure.  This mapping, however, is
less appealing from the computational complexity point of view since
it requires that one considers all possible complete assignments,
the number of which may be exponential in the size of the SCSP.

None of these two mappings are surjective, thus they cannot be used to
pass from a generic graphical game to an SCSP.  We also consider a
mapping which goes in this direction. This mapping creates a soft
constraint for each player, by looking at his neighbourhood.  We show
that this mapping defines a bijection between Pareto efficient joint
strategies and optimal solutions of the SCSP.

The study of the relations among preference models coming from different 
fields such as AI and game theory has only recently gained attention.
In \cite{gottlob} a mapping from the graphical games to 
hard CSPs has been defined, and it has been shown that 
the Nash equilibria of these games coincide with the solutions  
of the CSPs.
We can use this mapping, together with our mapping from the graphical
games to SCSPs, to identify the Pareto efficient Nash equilibria 
of the given game. In fact, these equilibria correspond to the optimal solutions 
of the SCSP obtained by joining the soft and hard constraints generated 
by the two mappings.
The mapping of \cite{gottlob}
leads to interesting results on the complexity of deciding 
whether a game has a pure Nash equilibrium or other kinds 
of desirable joint strategies.      

%In \cite{constNash} graphical games are considered where 
%there possibly are constraints on the outcomes, enforced 
%over the payoffs of the players, constraints on 
%the actions the players can choose and 
%objective functions on the global outcomes inducing 
%different optimality criteria over the equilibria.
%....DCOP     

In \cite{tambe} a mapping from distributed 
constraint optimization problems (DCOPs) to graphical games 
is introduced,
where the optimization criterion is to maximize the sum of 
utilities.
By using this mapping, it is shown that the optimal solutions 
of the given DCOP are Nash equilibria of the generated game. 
This result is in line with our finding regarding 
strictly monotonic SCSPs, which include the class of problems considered 
in \cite{tambe}.

\subsection{Organization of the paper}

The paper is organized as follows.  In Section \ref{back} we introduce
CP-nets, soft constraints, and strategic games.
Next, in Section
%\ref{sec:games}, we recall the basic concepts of strategic games and in Section 
\ref{sec:param} we introduce 
%our modification of 
%brent r1
a modification of the classical notion of 
strategic games considered in this paper. 
In Section \ref{sec:to-games} we show how to pass from
CP-nets to so defined strategic games, while in Section
\ref{sec:to-nets} we deal with the opposite direction.  

Then in Section \ref{sec:games-to-nets} we show how to apply
techniques developed in game theory to reason about CP-nets, while in
Section \ref{sec:nets-to-games} we study the other direction.  Next,
in Section \ref{sec:soft-to} and \ref{sec:to-soft}
we study the relationship between soft CSPs and
strategic games by relating optimal solutions of soft CSPs to Nash
equilibria and Pareto efficient joint strategies.  Finally, in Section
\ref{conc} we summarize the main contributions of the paper.

Preliminary results of this research were reported in \cite{ARV05} and 
\cite{ARV08}.

\section{Preliminaries}
\label{back}

In this section we recall the main notions regarding CP-nets,
soft constraints, and strategic games.

\subsection{CP-nets}

CP-nets~\cite{BBHP.UAI99,BBHP.journal} (for Conditional Preference nets)
are a graphical model for
compactly representing conditional and qualitative preference
relations. They exploit conditional preferential independence by
decomposing an agent's preferences via the \oldbfe{ceteris paribus} 
assumption. Informally, CP-nets are sets of \oldbfe{ceteris 
paribus (cp)\/} preference statements. For instance, the statement
{\em ``I prefer red wine to white wine if meat is served."} asserts
that, given two meals that differ {\em only} in the kind of wine
served {\em and} both containing meat, the meal with a red wine is
preferable to the meal with a white wine.
On the other hand, this statement does not order 
two meals with a different main course. 
Many users' preferences 
appear to be of this type.

CP-nets bear some similarity to Bayesian networks. Both utilize
directed graphs where each node stands for a domain variable, and
assume a set of \emph{features} (variables) $F = \{X_1,\ldots,X_n\}$
with the corresponding finite domains $\cal D$$(X_1),\ldots,$$\cal
D$$( X_n)$.  For each feature $X_i$, a user specifies a (possibly
empty) set of \oldbfe{parent} features $Pa(X_i)$ that can affect her
preferences over the values of $X_i$.  This defines a
%brent r2
directed graph, called  
\oldbfe{dependency graph}, in which each node $X_i$ has $Pa(X_i)$ as
its immediate predecessors.
A CP-net is said to be \oldbfe{acyclic} if its dependency graph 
does not contain cycles.  

Given this structural information, the user explicitly specifies her
preference over the values of $X_i$ for {\em each complete assignment}
on $Pa(X_i)$. In this paper this preference is assumed to take the
form of a linear order over $\cal D$$(X_i)$
\cite{BBHP.UAI99,BBHP.journal}.\footnote{In this we follow
  \cite{BBHP.journal}, where ties among values are initially allowed,
  (that is linear pre-orders are admitted) but in
  presentation only total orders are used. If ties are admitted, the notion of an optimal outcome of a CP-net
has to be appropriately modified.}  Each such specification
is called below a \oldbfe{preference statement} for the variable
$X_i$.  These conditional preferences over the values of $X_i$ are
captured by a {\em conditional preference table} which is annotated
with the node $X_i$ in the CP-net.  An \oldbfe{outcome} is an
assignment of values to the variables with each value taken from the
corresponding domain.

As an example, consider a CP-net 
whose features are $A$, $B$, $C$ and $D$, with binary domains
containing $f$ and $\overline{f}$ if $F$ is the name of the feature,
and with the following preference statements: 

$d: a \succ \overline{a}$, \ $\overline{d} : a \succ \overline{a}$,

$a : b \succ \overline{b}$, \ $\overline{a} : \overline{b} \succ b$, 

$b : c \succ \overline{c}$, \ $\overline{b} : \overline{c} \succ c$, 

$c : d \succ \overline{d}$, \ $\overline{c} : \overline{d} \succ d$.

\NI
Here the preference statement $d: a \succ \overline{a}$
states that $A=a$ is preferred to $A=\overline{a}$, given that $D = d$.
From the structure of these preference statements we see that
$Pa(A) = \C{D}, Pa(B) = \C{A}, Pa(C) = \C{B}, Pa(D) = \C{C}$
so the dependency graph is cyclic.

An \oldbfe{acyclic} CP-net is one in which the dependency graph is
acyclic. As an example, consider a CP-net
whose features and domains are as above
and with the following preference statements: 

$a \succ \overline{a}$, 

$b \succ \overline{b}$, 

$(a \wedge b) \vee (\overline{a} \wedge \overline{b}) : c \succ \overline{c}$, \ 
$(a \wedge \overline{b}) \vee (\overline{a} \wedge b) : \overline{c} \succ c$, 

$c: d \succ \overline{d}$, \ $\overline{c}: \overline{d} \succ d$.  

\NI
Here, the preference statement $a \succ \overline{a}$ represents the
unconditional preference for $A=a$ over $A=\overline{a}$.  
Also each preference statement for the variable $C$ 
is a actually an abbreviated version of two preference statements.
In this example we
have $Pa(A) = \ES, Pa(B) = \ES, Pa(C) = \C{A,B}, Pa(D) = \C{C}$.

%The semantics of CP-nets depends on the notion 
%of a \oldbfe{worsening flip}. 
A \oldbfe{worsening flip}
is a transition between two outcomes that consists
of a change in the value of a single variable to one which
is less preferred in the unique preference statement for that
variable.
By analogy we define an \oldbfe{improving flip}.
For example, in the acyclic CP-net 
%brent r2
described in the previous paragraph,
passing from $abcd$ to $ab\overline{c}d$ is a worsening flip since $c$
is better than $\overline{c}$ given $a$ and $b$.  We say that an
outcome $\alpha$ is \oldbfe{better} than the outcome $\beta$ 
(or, equivalently, $\beta$ is \oldbfe{worse} than $\alpha$),  
written as $\alpha \succ \beta$, iff there is a chain 
of worsening flips from $\alpha$ to $\beta$.
This definition induces a strict preorder over the outcomes.
In the acyclic CP-net 
%brent r2
described in the previous paragraph, 
the outcome 
$\overline{a}b\overline{c}\overline{d}$ is worse than $abcd$.

An \oldbfe{optimal} outcome is 
one for which no better outcome exists.
So an outcome is optimal iff no improving flip from it exists.
In general, a CP-net does not need to have an optimal outcome. As an
example consider two features $A$ and $B$ with the respective domains
$\C{a, \overline{a}}$ and $\C{b, \overline{b}}$ and the following
preference statements:

$a: b \succ \overline{b}$, \ $\overline{a}: \overline{b} \succ b$,

$b: \overline{a} \succ a$, \ $\overline{b}: a \succ \overline{a}$.

\NI
It is easy to see that then 

$
ab \succ a\overline{b} \succ \overline{a}\overline{b} \succ \overline{a}b \succ ab.
$

%\begin{changebar}
Finding optimal outcomes and testing for optimality 
is known to be NP-hard~\cite{BBHP.UAI99,BBHP.journal}.  
However, in acyclic CP-nets
there is a unique optimal outcome and it can be found in linear
time~\cite{BBHP.UAI99,BBHP.journal}. We simply sweep through the
CP-net, following the arrows in the dependency graph, assigning at
each step the most preferred value in the preference relation.  For
instance, in the CP-net
above, we would choose $A = a$ and $B=b$, then $C=c$ and then
$D=d$. The optimal outcome is therefore $abcd$.

Determining whether one outcome is better than another according to this
order (a so-called \emph{dominance query}) is also 
NP-hard even for acyclic CP-nets, see \cite{Domshlak:Brafman:kr02}.
Whilst tractable special cases exist, there are also acyclic CP-nets
in which there are exponentially long chains of worsening flips
between two outcomes~\cite{Domshlak:Brafman:kr02}.  

Hard constraints are enough to find optimal outcomes of a CP-net and
to test whether a CP-net has an optimal outcome.  In fact, given
a CP-net one can define a set of hard constraints (called
\oldbfe{optimality constraints}) such that their solutions are the
optimal outcomes of the CP-net, see \cite{dimo,aaai05}.
 
Indeed, take a CP-net $N$ and consider a linear order $\succ$ over
the elements of the domain of a variable $X$ used in a preference
statement for $X$. Let $\varphi$ be the disjunction of the
corresponding assignments used in the preference statements that use
$\succ$.  Then for each of such linear order $\succ$ the
corresponding optimality constraint is $\varphi \rightarrow X=a_j$,
where $a_j$ is the undominated element of $\succ$.  The optimality
constraints $opt(N)$ corresponding to $N$ consist of the entire set of
such optimality constraints, each for one such linear order
$\succ$.

For example, the preference statements $a \succ \overline{a}$ and 
$(a \wedge \overline{b}) \vee (\overline{a} \wedge b) : \overline{c} \succ c$
from the above CP-net map to the hard
constraints $A = a$ and 
$(A = a \wedge B = \overline{b}) \vee (A = \overline{a} \wedge B = b) 
\rightarrow C = \overline{c}$,
respectively.

It has been shown that an outcome is optimal in the strict preorder over the
outcomes induced by a CP-net $N$ iff it is a satisfying assignment for
$opt(N)$.

A CP-net is called \oldbfe{eligible} iff it has an optimal outcome.  Even if
the strict preorder induced by a CP-net has cycles, the CP-net may still be
useful if it is eligible.  All acyclic CP-nets are trivially eligible
as they have a unique optimal outcome.  We can thus test eligibility
of any (even cyclic) CP-net by testing the consistency of the
optimality constraints $opt(N)$. That is, a CP-net $N$ is eligible iff
$opt(N)$ is consistent.
%\end{changebar}

%\section{Strategic games with parametrized preferences}

%In this section we review the basic notions of strategic games and we 
%generalize them to a notion of games with parametrized preferences.

%%%%%%%%%%%%%%%%%qui mettere soft

\subsection{Soft constraints}
\label{subsec:soft}

Soft constraints, see e.g. \cite{jacm},
allow us to express constraints and preferences.
While constraints state which combinations of variable values are
acceptable, soft constraints (also called \emph{preferences})
allow for several levels of acceptance.
A technical way to describe soft constraints is via the use of 
an algebraic structure called a c-semiring.

%Soft constraints, see e.g. \cite{jacm}, model 
%brent r1 
%the preferences involved in a constraint problem
%with c-semirings.

A \oldbfe{c-semiring} is
a tuple $\langle A,+,\times,\0,\1 \rangle$, where:
\begin{itemize}
\item $A$ is a set, called the \oldbfe{carrier} of the semiring, and $\0, \1 \in A$;
\item $+$ is commutative, associative, idempotent,
$\0$ is its unit element, and $\1$ is its absorbing element;
\item $\times$ is associative, commutative,
distributes over $+$, $\1$  is its
unit element and $\0$ is its absorbing element.
\end{itemize}

Elements $\0$ and $\1$ represent, respectively, the highest and lowest
preference.  While the operator $\times$ is used to combine
preferences, the operator + induces a partial order on the carrier $A$
defined by
\[
\mbox{$a \leq b$ iff $a+b = b$.}
\]

Given a c-semiring $S = \langle A,+,\times,\0,\1 \rangle$,
and a set of variables $V$, each variable $x$ with a domain $D(x)$,
a \oldbfe{soft constraint}
is a pair $\langle def, con \rangle$, where
$con \subseteq V$  and $def: \times_{y \in con} D(y) \rightarrow A$.
So a constraint specifies a set of variables (the ones in $con$),
and assigns to each tuple of values from $\times_{y \in con} D(y)$,
the Cartesian product of the variable domains, an
element of the semiring carrier $A$.

A \oldbfe{soft constraint satisfaction problem} (in short,
a \oldbfe{soft CSP} or  \oldbfe{SCSP})
is a tuple $\langle C, V, D, S \rangle$ where $V$ is a set of variables,
with the corresponding set of domains $D$, $C$ is a set of soft constraints
over $V$ and $S$ is a c-semiring.
Given an SCSP a \oldbfe{solution} is an instantiation
of all the variables. The \oldbfe{preference} of a solution $s$ is
the combination by means of the $\times$
operator of all the preference levels given by the constraints to the
corresponding subtuples of the solution, or more formally,
\[
\times_{c \in C} def_c(s \downarrow_{con_c}),
\]
where $\times$ is the multiplicative operator of the semiring and
$def_c(s \downarrow_{con_c})$ is the preference associated by the
constraint $c$ to the projection of the solution $s$ on the variables
in $con_c$.

A solution is called \oldbfe{optimal} if there is no other
solution with a strictly higher preference.

Three widely used instances of SCSPs are:
\begin{itemize}
\item {\bf Classical CSPs} (in short {\bf CSPs}),
based on the c-semiring
$\langle \{0,1\},\lor,\land,$ $ 0, 1\rangle$. They model
 the customary CSPs in which tuples are either allowed or not. So CSPs can be
  seen as a special case of SCSPs.
\item {\bf Fuzzy CSPs}, based on the \oldbfe{fuzzy c-semiring}
  $\langle [0,1],max,min,0,1\rangle$.
   In such problems, preferences are the values in
  $[0,1]$, combined by taking the minimum and the goal is to
  maximize the minimum preference.
\item {\bf Weighted CSPs}, based on the \oldbfe{weighted c-semiring}
$\langle\Re_{+},min,+, \infty$ $, 0\rangle$.
Preferences are costs ranging over non-negative reals,
   which are aggregated using the sum. The goal is to minimize the
  total cost.
\end{itemize}

%\begin{itemize}
%\item a set of variables with finite domains,
%\item the \oldbfe{preference levels} that are taken from a set $A$,
%  over which a linear order\footnote{in general partial orders
%    are allowed} $\leq$ and an associative \oldbfe{combination
%    operator} $\times$ are defined,
%\item a set of \oldbfe{constraints}; each constraint involves a set of
%variables and assigns to each instantiation of its variables
%a preference level,
%\item the \oldbfe{preference} of solution (that is, an instantiation
%  of all the variables) is the combination by means of the $\times$
%  operator of all the preference levels given by the constraints to the
%  corresponding subtuples of the solution.
%\end{itemize}

%We talk then of a soft constraint satisfaction problem, in short a
%\oldbfe{soft CSP}.  The optimal solutions to a soft CSP are then the
%ones with the $\leq$-maximal preference.

%A widely used instance of this formalism is the class of \oldbfe{fuzzy
%  constraints}, see \cite{fuzzy1} and \cite{ruttkay-fuzzy}, where the
%set of preference levels $A$ is the real interval $[0,1]$, $\leq$ is
%the usual inequality order over the reals, and the combination
%operator is the \emph{min} operation.  In other words, in the fuzzy
%constraint satisfaction problems (\oldbfe{fuzzy CSPs}) the
%preference of a solution is the minimum of the preferences of
%all the subtuples selected by the constraints.

A simple example of a fuzzy CSP is the following one:
\begin{itemize}
\item three variables: $x$, $y$, and $z$, each with the domain $\{a,b\}$;
\item two constraints: $C_{xy}$ (over $x$ and $y$) and $C_{yz}$ (over
  $y$ and $z$) defined by:

$C_{xy} := \{(aa, 0.4), (ab, 0.1), (ba, 0.3), (bb, 0.5)\}$,

$C_{yz} := \{(aa, 0.4), (ab, 0.3), (ba, 0.1), (bb, 0.5)\}$.
\end{itemize}
The unique optimal solution of this problem is $bbb$ (an
abbreviation for $x=y=z=b$). Its
preference is $0.5$.

\subsection{Strategic games}
\label{sec:games}

Let us recall now the notion of a strategic game, see, e.g.,
\cite{Mye91}.  A strategic game for a set $N = \{1, \LL, n\}$ of $n$ players ($n > 1$) is a
tuple
\[
(S_1, \LL, S_n, p_1, \LL, p_n),
\] 
where for each $i \in [1..n]$

\begin{itemize}
\item $S_i$ is the non-empty set of \oldbfe{strategies} 
available to player $i$,

\item $p_i$ is the \oldbfe{payoff function} for the  player $i$, so
$
p_i : S_1 \times \LL \times S_n \myra \cal{R},
$
where $\cal{R}$ is the set of real numbers.
\end{itemize}

Given a sequence of non-empty
sets $S_1, \LL, S_n$ and $s \in S_1 \times \LL \times S_n$ we denote
the $i$th element of $s$ by $s_i$, abbreviate $N \setminus \{i\}$ to $-i$,
and use the following standard
notation of game theory, where $i \in [1..n]$ and $I := i_1, \LL, i_k$
is a subsequence of $1, \LL, n$:

\begin{itemize}
\item $s_{I} := (s_{i_1}, \LL, s_{i_k})$,

\item $(s'_i, s_{-i}) := (s_1, \LL, s_{i-1}, s'_i, s_{i+1}, \LL, s_n)$, where
we assume that $s'_i \in S_i$,

\item $S_{I} := S_{i_1} \times  \LL \times  S_{i_k}$.
\end{itemize}

A joint strategy $s$ is called  

\begin{itemize}
\item a \oldbfe{pure Nash equilibrium} (from now on, simply \oldbfe{Nash equilibrium})
if 
\begin{equation}
  \label{eq:nash}
p_i(s) \geq p_i(s'_i, s_{-i})  
\end{equation}
for all $i \in [1..n]$ and all $s'_i \in S_i$,

\item \oldbfe{Pareto efficient}
if for no joint strategy $s'$,
$
p_i(s') \geq p_i(s)
$
for all $i \in [1..n]$ and 
$
p_i(s') > p_i(s)
$
for some $i \in [1..n]$.
\end{itemize}

Pareto efficiency can be alternatively defined by considering
the following strict \oldbfe{Pareto order} $<_P$ on the $n$-tuples of reals:
\[
\mbox{$(a_1, \LL, a_n) <_P (b_1, \LL, b_n)$ iff $ \fa i \in [1..n] \ a_i \leq b_i$ 
and $\te i \in [1..n] \ a_i < b_i$.} 
\]
Then a joint strategy $s$ is Pareto efficient iff
the $n$-tuple $(p_1(s), \LL, p_n(s))$ is a maximal element in the $<_P$
order on such $n$-tuples of reals.

To clarify these notions consider the classical Prisoner's Dilemma
game represented by the following bimatrix representing the payoffs to
both players:
\begin{center}
\begin{game}{2}{2}
       & $C_2$    & $N_2$\\
$C_1$   &$3,3$   &$0,4$\\
$N_1$   &$4,0$   &$1,1$
\end{game}
\end{center}

Each player $i$ represents a prisoner, who has two strategies, $C_i$
(cooperate) and $N_i$ (not cooperate).  Table entries represent
payoffs for the players (where the first component is the payoff of
player 1 and the second one that of player 2).

The two prisoners gain when both cooperate (with a profit of 3 each).
However, if only one of them cooperates, the other one
will gain more (with a profit of 4). If none of them
cooperates, both gain very little (a profit of 1 each), but more than the
"cheated" prisoner whose cooperation is not returned (that is, 0).

Here the unique Nash equilibrium is $(N_1,
N_2)$, while the other three joint strategies $(C_1, C_2), \ (C_1,
N_2)$ and $(N_1, C_2)$ are Pareto efficient.

A natural modification of the concept of strategic games, called
graphical games, was proposed in \cite{KLS01}. These games
stress the locality in taking decision.  In a graphical game the
payoff of each player depends only on the strategies of its neighbours
in a given in advance graph structure over the set of players.

Formally, a \oldbfe{graphical game} for $n$ players with the corresponding
strategy sets $S_1, \LL, S_n$ is defined by assuming a neighbour
function \emph{neigh} that given a player $i$ yields its set of
neighbours $\emph{neigh}(i)$. The payoff for player $i$ is then a
function $p_i$ from $\times_{j \in \emph{neigh}(i) \cup \{i\}} S_j$ to $\cal{R}$. 

By using the canonic extensions of these payoff functions to the
Cartesian product of all strategy sets one can then extend the
previously introduced concepts, notably that of a Nash equilibrium, to
the graphical games.  Further, when all pairs of players are
neighbours, a graphical game reduces to a strategic game.

\section{Strategic games with parametrized preferences}
\label{sec:param}

In game theory it is customary to study strategic games defined as
above, in quantitative terms.  A notable exception is \cite{OR94} in
which instead of payoff functions the linear quasi-orders on the
sets of joint strategies are used.

For our purposes we need a different approach.  To define it we first
introduce the concept of a \oldbfe{preference} on a set $A$ which in
this paper denotes a strict linear order on $A$.  We then assume
that each player has to his disposal a preference relation $\succ
\hspace{-1.5mm}(s_{-i})$ on his set of strategies \emph{parametrized}
by a joint strategy $s_{-i}$ of his opponents.  So in our approach
\begin{itemize}

\item  for each $i \in [1..n]$ player $i$ has
a finite, non-empty, set $S_i$ of strategies available to him,

\item for each $i \in [1..n]$ and $s_{-i} \in S_{-i}$
player $i$ has a preference relation $\succ\hspace{-1.5mm}(s_{-i})$
on his set of strategies $S_i$. 
\end{itemize}

In what follows such a \oldbfe{strategic game with parametrized
  preferences} (in short a \oldbfe{game with parametrized
  preferences}, or just a \oldbfe{game}) for $n$ players is
represented by a tuple
\[
(S_1, \LL, S_n, {\succ}(s_{-1}), \LL, {\succ}(s_{-n})),
\] 
where each $s_{-i}$ ranges over $S_{-i}$.

It is straightforward to transfer to the case of games with
parametrized preferences the basic notions concerning strategic games.
In particular the following notions will be of importance for us
(for the original definitions see, e.g., \cite{OR94}), where
$G$ is a game with parametrized preferences specified as above:

  \begin{itemize}
  \item A strategy $s_i$ is a \oldbfe{best response} for
    player $i$ to a joint strategy $s_{-i}$ of his opponents if
$s_i \succ\hspace{-1.5mm}(s_{-i}) \ s'_i$,
for all $s'_i \neq s_i$. 

\item A strategy $s_i$ is \oldbfe{never a best response}
  for player $i$ if it is not a best response to any joint 
  strategy $s_{-i}$ of his opponents.  

\item A strategy $s'_i$ is
 \oldbfe{strictly dominated} by a strategy $s_i$ if
$s_i \succ\hspace{-1.5mm}(s_{-i}) \ s'_i$,
for all $s_{-i} \in S_{-i}$.
\end{itemize}

So according to this terminology a joint strategy $s$ is a \oldbfe{Nash
equilibrium} of $G$ iff each $s_i$ is a best response to $s_{-i}$.
Note, however, that in our setup the underlying preferences are
strict, so the above notions of a best response and Nash equilibrium
correspond in the customary setting of strategic games to the notions
of a unique best response and a strict Nash equilibrium. In
particular, note that $s$ is a Nash equilibrium of $G$ iff for all $i
\in [1..n]$ and all $s'_i \neq s_i$
\[
s_i \succ \hspace{-1.5mm}(s_{-i}) \ s'_i,
\]
because to each joint strategy $s_{-i}$ a unique best response exists.

To clarify these definitions let us return to the above example of the
strategic game that models the Prisoner's Dilemma.
To view this game as a game with parametrized preferences we abstract from
the numerical values and simply
stipulate that
\II

$\succ \hspace{-1.5mm}(C_2) := \ N_1 \succ C_1$, \ 
$\succ \hspace{-1.5mm}(N_2) := \ N_1 \succ C_1$,

$\succ \hspace{-1.5mm}(C_1) := \ N_2 \succ C_2$, \ 
$\succ \hspace{-1.5mm}(N_1) := \ N_2 \succ C_2$.
\II

\NI
These orders reflect the fact that for each strategy of the
opponent each player considers his `not cooperate' strategy better
than his `cooperate' strategy.

%\begin{changebar}
It is easy to check that:

\begin{itemize}

\item for each player $i$ the strategy $C_i$ is strictly dominated by $N_i$
(since $N_i \succ \hspace{-1.5mm}(C_{3-i})  C_i$ and
$N_i \succ \hspace{-1.5mm}(N_{3-i}) C_i$),

\item for each player $i$ the strategy $N_i$ is a best response to the strategy
$N_{3-i}$ of his opponent,

\item (as a result) $(N_1, N_2)$ is a unique Nash equilibrium of this game with 
parametrized preferences.
\end{itemize}

The framework of the games with parametrized preferences allows us to
discuss only some aspects of the customary strategic games. In
particular it does not allow us to introduce the notion of a mixed
strategy, since the outcomes of playing different strategies by a
player, given the joint strategy chosen by the opponents, cannot be
aggregated.  Also the notion of a Pareto efficient outcome does not
have a counterpart in this framework because in general two joint
strategies cannot be compared.  For example, in the above modelling of
the Prisoner's Dilemma game we cannot compare the joint strategies
$(N_1, N_2)$ and $(C_1, C_2)$.

In the field of strategic games two techniques of reducing a game
have been considered --- by means of iterated elimination of strategies
strictly dominated by a mixed strategy or of iterated elimination of
never best responses to a mixed strategy (see, e.g., \cite{OR94}.)
These techniques can be easily transferred to the games with parametrized
preferences provided we limit ourselves to strict dominance by a pure strategy
and never best responses to a pure strategy.

First, given such a game
\[
G := (S_1, \LL, S_n, {\succ}(s_{-1}), \LL, {\succ}(s_{-n})),
\] 
where each $s_{-i}$ ranges over $S_{-i}$, and
sets of strategies $S'_1, \LL, S'_n$ such that $S'_i \sse S_i$ for $i \in
[1..n]$, we say that 
\[
G' := (S'_1, \LL, S'_n, {\succ}(s_{-1}), \LL, {\succ}(s_{-n})),
\] 
where each $s_{-i}$ now ranges over $S'_{-i}$,
is a \oldbfe{subgame} of $G$, and identify in the context of $G'$ each
preference relation $\succ\hspace{-1.5mm}(s_{-i})$ with its restriction to $S'_i$.

We then introduce the following two notions of reduction between a game
\[G := (S_1, \LL, S_n, {\succ}(s_{-1}), \LL, {\succ}(s_{-n})),
\]
where each $s_{-i}$ ranges over $S_{-i}$ and its subgame
\[
G' := (S'_1, \LL, S'_n, {\succ}(s_{-1}), \LL, {\succ}(s_{-n})),
\]
where each $s_{-i}$ ranges over $S'_{-i}$:

\begin{itemize}
\item 
$
G \myra_{\hspace{-1mm} NBR \: } G'
$

when $G \neq G'$ and for all $i \in [1..n]$ 
each $s_i \in S_i \setminus S'_i$ is never a best response for player $i$ in $G$,

\item 
$
G \myra_{\hspace{-1mm} S \: } G'
$

when $G \neq G'$ and for all $i \in [1..n]$ 
each $s'_i \in S_i \setminus S'_i$ is strictly dominated in $G$ by some $s_i \in S_i$.
\end{itemize}

In the literature it is customary to consider more specific reduction
relations in which, respectively, \emph{all} never best responses or
\emph{all} strictly dominated strategies are eliminated. The advantage
of using the above versions is that we can prove the relevant property
of both reductions by just one simple lemma, since by definition a
strictly dominated strategy is never a best response and consequently
$G \myra_{\hspace{-1mm} S} G'$ implies $G \myra_{\hspace{-1mm} NBR}
G'$.

\begin{lemma}
\label{lem:nbr}

Suppose that $G \myra_{\hspace{-1mm} NBR} G'$.
Then $s$ is a Nash equilibrium of $G$ iff it is a Nash equilibrium of $G'$.
\end{lemma}

%\begin{changebar}
\Proof 
$(\Ra)$
By definition each $s_i$ is a best response to $s_{-i}$ to $G$.
So no $s_i$ is eliminated in the reduction of $G$ to $G'$.
\II

\NI
$(\La)$
Suppose $s$ is not a Nash equilibrium of $G$. So
some $s_i$ is not a best response to $s_{-i}$ in $G$. Let 
$s'_i$ be a best response to $s_{-i}$ in $G$.
($s'_i$ exists 
since $\succ\hspace{-1.5mm}(s_{-i})$ is a linear order.)

So $s'_i$ is not eliminated in the reduction of $G$ to $G'$ and $s'_i$
is a best response to $s_{-i}$ in $G'$.  But this contradicts the fact
that $s$ is a Nash equilibrium of $G'$.  
\HB
%\end{changebar}

\begin{theorem}
\label{thm:ienbr}

Suppose that $G \myra^{*}_{\hspace{-1mm} NBR} G'$, i.e., $G'$ is
obtained by an iterated elimination of never best responses from the
game $G$.  

  \begin{enumerate} \smallromani
  \item Then $s$ is a Nash equilibrium of $G$ iff it is a Nash equilibrium of $G'$.
    
  \item If each player in $G'$ has just one strategy, 
then the resulting joint strategy is a
    unique Nash equilibrium of $G$.  
  \end{enumerate}
\end{theorem}
%\begin{changebar}
\Proof

\NI
$(i)$ By the repeated application of 
Lemma \ref{lem:nbr}. 
\II

\NI
$(ii)$ It suffices to note that $(s_1, \LL s_n)$ is a unique Nash equilibrium of
the game in which each player $i$ has just one strategy,
$s_i$.
\HB
%\end{changebar}
\VV

The above theorem allows us to reduce a game without affecting its
(possibly empty) set of Nash equilibria or even, occasionally, to find
its unique Nash equilibrium.  In the latter case one says that the
original game was \oldbfe{solved} by an iterated elimination of never
best responses (or of strictly dominated strategies).

As an example let us return to the Prisoner's Dilemma game with
para- metrized preferences defined above.  In this game each strategy
$C_i$ is strictly dominated by $N_i$, so the game can be solved by
either reducing it in two steps (by removing in each step one $C_i$
strategy) or in one step (by removing both $C_i$ strategies) to a game
in which each player $i$ has exactly one strategy, $N_i$.

Finally, let us mention that \cite{GKZ90} and \cite{Ste90} proved that
all iterated eliminations of strictly dominated strategies yield the
same final outcome.  An analogous result for the iterated elimination of
never best responses was established in \cite{Apt05}.
Both results carry over to our framework of
games with parametrized preferences by a direct modification of
the proofs.

\section{From CP-nets to strategic games}
\label{sec:to-games}

Consider now a CP-net with the set of variables $\C{X_1, \LL, X_n}$
with the corresponding finite domains ${\cal D}(X_1), \LL, {\cal D}(X_n)$.
We write each preference statement for the variable $X_i$ as 
$X_I = a_I : \ \succ_i$,
where for the subsequence $I = i_1, \LL, i_k$ of $1, \LL, n$:
\begin{itemize}
\item $Pa(X_i) = \C{X_{i_1}, \LL, X_{i_k}}$,
\item $X_I = a_I$ is an abbreviation for
$X_{i_1} = a_{i_1} \A \LL \A X_{i_k} = a_{i_k}$,
\item $\succ_i$ is a preference over ${\cal D}(X_i)$.
\end{itemize}
We also abbreviate ${\cal D}(X_{i_1}) \times \LL \times {\cal D}(X_{i_k})$
to ${\cal D}(X_I)$.

By definition, the preference statements for a variable $X_i$
are exactly all statements of the form 
$X_I = a_I: \ \succ\hspace{-1.5mm}(a_I)$, 
where $a_I$ ranges over ${\cal D}(X_I)$ and 
$\succ\hspace{-1.5mm}(a_I)$ is a preference on 
${\cal D}(X_i)$ that depends on $a_I$.

We now associate with each 
CP-net $N$ a game ${\cal G}(N)$ with parametrized preferences
as follows:
\begin{itemize}
\item each variable $X_i$ corresponds to a player $i$,
\item the strategies of player $i$ are the elements of the domain ${\cal D}(X_i)$
of $X_i$.
\end{itemize}

To define the parametrized preferences, consider a player $i$. Suppose
$Pa(X_i) = \C{X_{i_1}, \LL, X_{i_k}}$ and let $I := i_1, \LL, i_k$. So
$I$ is a subsequence of $1, \LL, i-1, i+1, \LL, n$ and consequently each
joint strategy $a_{-i}$ of the opponents of player $i$ uniquely determines
a sequence $a_{I}$. Given now an arbitrary 
$a_{-i}$ we associate with it
the preference relation 
$\succ\hspace{-1.5mm}(a_{I})$ on ${\cal D}(X_i)$ where
$X_I = a_I: \  \succ\hspace{-1.5mm}(a_{I})$ is the 
unique preference statement
for $X_i$ determined by $a_I$.

%brent r2
In words, 
the preference of a player $i$ over his strategies,
assuming the joint strategy $a_{-i}$ of its opponents, 
coincides with the preference given by the CP-net over the domain 
of $X_i$, assuming the
assignment to its parents $a_{I}$ (which coincides with the projection
of $a_{-i}$ over $I$).  This completes the definition of ${\cal
  G}(N)$.

As an example  consider the first CP-net of Section \ref{back}.
The corresponding game has four players $A$, $B$, $C$, $D$, each with
two strategies indicated with $f$, $\bar{f}$ for player $F$.
The preference of each player on his strategies will depend only on the
strategies chosen by the players which correspond to his parents in the
CP-net. 
Consider for example player $B$. His preference
over his strategies $b$ and $\bar{b}$,
given the joint strategy of his opponents $s_{-B}=dac$, is 
$b \succ \bar{b}$. Notice that, for example, the
same order holds for the opponents joint strategy
$s_{-B}=\bar{d}a\bar{c}$, since the strategy chosen by the only player
corresponding to his parent, $A$, has not changed.     
 
We have then the following result.

\begin{theorem} \label{thm:G(N)}
An outcome of a CP-net $N$ is optimal iff it is 
a Nash equilibrium of the game ${\cal G}(N)$.
\end{theorem}
%\begin{changebar}
\Proof
$(\Ra)$
Take an optimal outcome $o$ of $N$.
Consider a player $i$ in the game ${\cal G}(N)$
and the corresponding variable $X_i$ of $N$.
Suppose $Pa(X_i) = \C{X_{i_1}, \LL, X_{i_k}}$. Let $I := i_1, \LL, i_k$,
and let $X_I = o_I: \  \succ\hspace{-1.5mm}(o_{I})$ be the corresponding
preference statement for $X_i$.
By definition there is no improving flip from $o$ to another outcome,
so $o_i$ is the maximal element in the order $\succ\hspace{-1.5mm}(o_{I})$.

By the construction of the game ${\cal G}(N)$, each outcome in $N$ is
a joint strategy in ${\cal G}(N)$. Also, two outcomes are one flip
away iff the corresponding joint strategies differ only in a strategy
of one player.  Given the joint strategy $o$ considered above, we thus
have that, if we modify the strategy of player $i$, while leaving the
strategies of the other players unchanged, this change is worsening in
$\succ\hspace{-1.5mm}(o_{-i})$, since
$\succ\hspace{-1.5mm}(o_{-i})$ coincides with
$\succ\hspace{-1.5mm}(o_I)$. So by definition $o$ is a Nash
equilibrium of ${\cal G}(N)$.
\II

\NI
$(\La)$
Take a Nash equilibrium $s$ of the game ${\cal G}(N)$.
Consider a  variable $X_i$ of $N$.
Suppose $Pa(X_i) = \C{X_{i_1}, \LL, X_{i_k}}$. Let $I := i_1, \LL, i_k$,
and let $X_I = s_I: \  \succ\hspace{-1.5mm}(s_{I})$ be the corresponding
preference statement for $X_i$.

By definition for every strategy $s'_i \neq s_i$ of player $i$,
we have $s_i \succ\hspace{-1.5mm}(s_{-i}) \ s'_i$, so
$s_i \succ\hspace{-1.5mm}(s_{I}) \ s'_i$ since
$\succ\hspace{-1.5mm}(s_{-i})$ coincides with
$\succ\hspace{-1.5mm}(s_I)$. So by definition $s$ is
an optimal outcome for $N$.
\HB

\section{From strategic games to CP-nets}
\label{sec:to-nets}

We now associate with each game $G$ with parametrized preferences 
a CP-net ${\cal N}(G)$ as follows:

\begin{itemize}
\item each variable $X_i$ corresponds to a player $i$,

\item the domain ${\cal D}(X_i)$ of the variable $X_i$ consists of
  the set of strategies of player $i$,
\item we stipulate that $Pa(X_i) = \C{X_{1}, X_{i-1}, \LL, X_{i+1}, \LL, X_n}$,
where $n$ is the number of players in $G$.
\end{itemize}

Next, for each joint strategy $s_{-i}$ of 
the opponents of player $i$ we take
the preference statement
$X_{-i} = s_{-i}: \  \succ\hspace{-1.5mm}(s_{-i})$,
where $\succ\hspace{-1.5mm}(s_{-i})$ is the preference relation on the set of strategies
of player $i$ associated with $s_{-i}$.

This completes the definition of ${\cal N}(G)$.  As an example of this
construction let us return to the Prisoner's Dilemma game with
parametrized preferences from Section \ref{sec:games}.  
In the corresponding CP-net we have then
two variables $X_1$ and $X_2$ corresponding to players 1 and 2, with
the respective domains $\C{C_1, N_1}$ and $\C{C_2, N_2}$.  To explain how
each parametrized preference translates to a preference statement take
for example $\succ \hspace{-1.5mm}(C_2) := \ N_1 \succ C_1$. It
translates to $X_2 = C_2 : \ N_1 \succ C_1$.

\NI
We have now the following counterpart of Theorem \ref{thm:G(N)}.

\begin{proposition} \label{thm:N(G)}
A joint strategy is a Nash equilibrium of the game $G$ iff it is an optimal
outcome of the CP-net ${\cal N}(G)$.
\end{proposition}
%\begin{changebar}
\Proof
It suffices to notice that  ${\cal G}({\cal N}(G)) = G$ and use
Theorem \ref{thm:G(N)}.
\HB
\VV

The disadvantage of the above construction of the CP-net ${\cal N}(G)$ from a
game $G$ is that it always produces a CP-net in which all sets of
parent features are of size $n-1$ where $n$ is the number of features
of the CP-net.  This can be rectified by reducing each set of parent
features to a minimal one as follows.

Given a CP-net $N$, consider a variable $X_i$ with the parents 
$Pa(X_i)$, and take a variable $Y \in Pa(X_i)$. 
Suppose that for all assignments $a$ to
$Pa(X) - \{Y\}$ and any two values $y_1,y_2 \in {\cal D}(Y)$, 
the orders  $\succ\hspace{-1.5mm}(a,y_1)$ and $\succ\hspace{-1.5mm}(a,y_2)$
on ${\cal D}(X_i)$ coincide.

We say then that $Y$ is \emph{redundant} in the set of parents of $X_i$.
It is easy to see that by removing all redundant variables 
from the set of parents of $X_i$ and 
by modifying the corresponding preference statements for $X_i$ accordingly, 
the strict preorder $\succ$ over the outcomes of the CP-nets is not changed.

Given a CP-net, if for all its variable $X_i$ the set  $Pa(X_i)$ 
does not contain any redundant variable, we say that the CP-net is 
\oldbfe{reduced}.

By iterating the above construction every CP-net can be transformed to
a reduced CP-net. 
As an example consider a CP-net with three features,
$X, Y$ and $Z$, with the respective domains $\C{a_1, a_2}, \C{b_1,
 b_2}$ and $\C{c_1, c_2}$.  Suppose now that $Pa(X) = Pa(Y) = \ES,
Pa(Z) = \C{X, Y}$ and that

$\succ\hspace{-1.5mm}(a_1, b_1) = \ \succ\hspace{-1.5mm}(a_2, b_1)$, \ 
$\succ\hspace{-1.5mm}(a_1, b_2) = \ \succ\hspace{-1.5mm}(a_2, b_2)$, 

$\succ\hspace{-1.5mm}(a_1, b_1) = \ \succ\hspace{-1.5mm}(a_1, b_2)$, \ 
$\succ\hspace{-1.5mm}(a_2, b_1) = \ \succ\hspace{-1.5mm}(a_2, b_2)$.

\NI
Then both $X$ and $Y$ are redundant in the set of parents of $Z$, 
so we can reduce the CP-net
by 
reducing $Pa(Z)$ to $\emptyset$.
$Z$ becomes an independent variable in the reduced CP-net with the
order over 
its domain which coincides with $\succ\hspace{-1.5mm}(a_1, b_1)$
(which is the same as the other three orders on the domain of $Z$).

%Brent: I removed what follows
% to $\C{Y}$ or to $\C{X}$.
%In the first case the new preference statements for $Z$ are:

%$Y = b_1 : \ \succ\hspace{-1.5mm}(b_1)$, \ 
%$Y = b_2 : \ \succ\hspace{-1.5mm}(b_2)$,
%
%\NI
%and in the latter case
%
%$X = a_1 : \ \succ\hspace{-1.5mm}(a_1)$, \ 
%$X = a_2 : \ \succ\hspace{-1.5mm}(a_2)$.

%Suppose now that

%$\succ\hspace{-1.5mm}(a_1) \neq \ \succ\hspace{-1.5mm}(a_2)$, \ 
%$\succ\hspace{-1.5mm}(b_1) \neq \ \succ\hspace{-1.5mm}(b_2)$.

%\NI
%Then each of the above two reductions yields a different reduced CP-net.

\begin{figure}[htbp]
\begin{center} \ \setlength{\epsfxsize}{3.3in}
\epsfbox{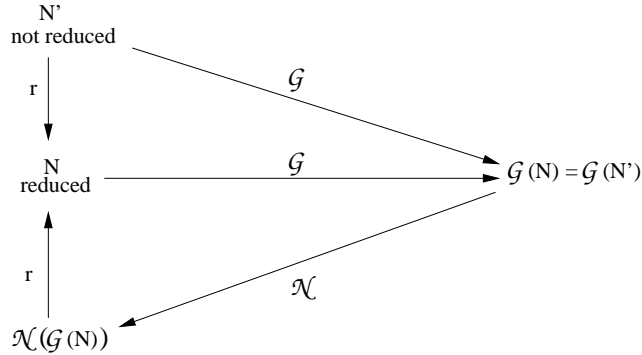}
\end{center}
\caption{Relation between a CP-net N, its reduced form and
  corresponding games}\label{teo4}
\end{figure}

In what follows for a CP-net $N$ we denote by 
$r(N)$ the corresponding 
%Apt:set of 
%Removed by Brent
reduced CP-net.
The following result, depicted in Figure \ref{teo4}, 
summarizes the relevant properties of $r(N)$ and
relates it to the constructions of ${\cal G}(N)$ and ${\cal N}(G)$.

\begin{proposition} \label{thm:reduced}
\mbox{} \vspace{-3mm}

\begin{enumerate} \smallromani

\item Each CP-net $N$ and its reduced form 
$r(N)$
have the same order $\succ$ over the outcomes. 

\item For each CP-net $N$ and its reduced form $r(N)$ we have
${\cal G}(N) = {\cal G}(r(N))$. 

\item Each reduced CP-net $N$ is a 
reduced CP-net corresponding to the game ${\cal G}(N)$.
Formally: $N = r({\cal N}({\cal G}(N)))$.
\end{enumerate} 
\end{proposition}
%\begin{changebar}
\Proof 
%Apt:TO REVISE
%It should be fine now we say that r(N) is unique

Statements (i) and (ii) easily follow from 
the definition of function $r$ and from 
from the construction of the game 
corresponding to a CP net.
We will thus write explicitly only the proof
of statement (iii).

%\NI 
%$(i)$ For every variable $X_i$ of $N$ with $Pa(X_i) = \C{X_{i_1},
%  \LL, X_{i_k}}$ and $I := i_1, \LL, i_k$ let $J$ be the
%subsequence of $I$ corresponding to the parents of $X_i$ in $r(N)$.
%By the definition of $r(N)$ for all $a_I \in {\cal D}(X_I)$ the
%orders $\succ\hspace{-1.5mm}(a_{I})$ and
%$\succ\hspace{-1.5mm}(a_{J})$ coincide, where $X_I = a_I: \ 
%\succ\hspace{-1.5mm}(a_{I})$ is the unique preference statement for
%$X_i$ in $N$ determined by $a_I$ and similarly with
%$\succ\hspace{-1.5mm}(a_{J})$.

%Consequently a flip is improving in $N$ iff it is improving in $r(N)$.

%\II

%\NI
%$(ii)$
%It follows directly from the construction of the game 
%corresponding to a CP net, since the preference of player 
%$i$ over its strategies depends on the strategies 
%of all the other opponents.
%\II

\NI 
$(iii)$ Given a reduced CP-net $N$, consider the CP-net ${\cal
  N}({\cal G}(N))$.  For each variable $X_i$, $Pa(X_i)$ in $N$ is a
subset of $Pa(X_i)$ in ${\cal N}({\cal G}(N))$, which is the set of
all variables except $X_i$. However, by the construction of the game
corresponding to a CP-net and of the CP-net corresponding to a game,
in each conditional preference table, if the assignments to the common
parents are the same, the preference orders over $X_i$ are the
same.

Let us now reduce ${\cal N}({\cal G}(N))$ to obtain $N' = r({\cal
  N}({\cal G}(N)))$.  Then $Pa(X_i)$ in $N'$ coincides with $Pa(X_i)$
in $N$.  Indeed, suppose there is a parent of $X_i$ in $N$ which is
not in $N'$. Since $N$ is reduced, such a parent is not redundant in
$N$. Thus the reduction $r$, when applied to ${\cal N}({\cal G}(N))$,
does not remove this parent since the orders in the conditional
preference tables of $N$ and ${\cal N}({\cal G}(N))$ are the same.

Further, suppose there is a parent of $X_i$ in $N'$ which is not in
$N$. Since $N'$ is reduced, such a parent is not redundant in $N'$.
Thus it is also not redundant in ${\cal N}({\cal G}(N))$.  By the
construction of ${\cal N}({\cal G}(N))$, this parent is not redundant in $N$
either.  
\HB
%\end{changebar}
\VV

Part $(i)$ states that the reduction procedure $r$
preserves the order over the outcomes.
Part $(ii)$ states that the construction of a game corresponding to 
a CP-net does not depend on the redundancy of the given CP-net.
Finally, part $(iii)$
states that the reduced CP-net $N$ can be obtained `back' from the game 
${\cal G}(N)$.

\section{Game-theoretic techniques in CP-nets}
\label{sec:games-to-nets}

Thanks to the established connections between CP-nets and games with
parametrized preferences, we can now transfer to CP-nets the
techniques of iterated elimination of strictly dominated strategies or
of never best responses considered in Section \ref{sec:games}.
To introduce them in the context of CP-nets
consider a CP-net $N$
with the set of variables $\C{X_1, \LL, X_n}$
with the corresponding finite domains ${\cal D}(X_1), \LL, {\cal D}(X_n)$.

\begin{itemize}

\item We say that an element $d_i$ from 
the domain $\cal D$$(X_i)$ of the variable $X_i$
is a \oldbfe{best response} to a preference statement
\[
X_I = a_I : \ \succ_i
\]
for $X_i$ 
if $d_i \succ_i d'_i$ 
for all $d'_i \in {\cal D}(X_i)$ such that $d_i \neq d'_i$.

\item We say that an element $d_i$ from the domain of the variable $X_i$
is a \oldbfe{never a best response}
if it is not a best response to any preference statement for $X_i$.

\item Given two elements $d_i, d'_i$ from the domain $\cal D$$(X_i)$
  of the variable $X_i$ we say that $d'_i$ is \oldbfe{strictly
    dominated} by $d_i$ if for all preference statements $X_I = a_I :
  \ \succ_i$ for $X_i$ we have
\[
d_i \succ_i d'_i.
\] 

\end{itemize}

By a \oldbfe{subnet} of a CP-net $N$ we mean a CP-net obtained from $N$
by removing some elements from some variable domains followed by the removal
of all preference statements that refer to a removed element.

Then we introduce the following 
relation between a CP-net $N$ and its subnet $N'$:
\[
N \myra_{\hspace{-1mm} NBR \: } N'
\]
when $N \neq N'$ and for each variable $X_i$ 
each removed element from the domain of $X_i$ is never a best response
in $N$, and also introduce an analogous relation $N \myra_{\hspace{-1mm} S \: } N'$
for the case of strictly dominated elements.
Since each strictly dominated element is never a best response, 
$N \myra_{\hspace{-1mm} S \: } N'$ implies
$N \myra_{\hspace{-1mm} NBR \: } N'$.

The following counterpart of Theorem \ref{thm:ienbr} then holds.

\begin{theorem}
\label{thm:nets-ienbr}
Suppose that $N \myra^{*}_{\hspace{-1mm} NBR} N'$, i.e., the CP-net $N'$ is
obtained by an iterated elimination of never best responses from the
CP-net $N$.  

  \begin{enumerate} \smallromani
  \item Then $s$ is an optimal outcome of $N$ iff it is an optimal outcome of $N'$.
    
  \item If each variable in $N'$ has a singleton domain, then the resulting outcome is a 
    unique optimal outcome of $N$.  

%  \item All iterated eliminations of never best responses from the
%CP-net $N$ yield the same final outcome. 
\HB
  \end{enumerate}
\end{theorem}

To illustrate the use of this theorem reconsider the first CP-net from
Section \ref{back}, i.e., the one with the preference statements

$d: a \succ \overline{a}$, \ $\overline{d} : a \succ \overline{a}$,

$a : b \succ \overline{b}$, \ $\overline{a} : \overline{b} \succ b$, 

$b : c \succ \overline{c}$, \ $\overline{b} : \overline{c} \succ c$, 

$c : d \succ \overline{d}$, \ $\overline{c} : \overline{d} \succ d$.

\NI
Denote it by $N$.

We can reason about it using the iterated elimination of strictly dominated strategies
(which coincides here with the iterated elimination  of never best responses, since each
domain has exactly two elements).

We have the following chain of reductions:
\[
N \myra_{\hspace{-1mm} S} N_1 \myra_{\hspace{-1mm} S} N_2 \myra_{\hspace{-1mm} S} N_3 \myra_{\hspace{-1mm} S} N_4,
\]
where

\begin{itemize}
\item $N_1$ results from $N$ by removing $\overline{a}$ (from the domain of $A$) and the 
preference statements $d: a \succ \overline{a}$, \ $\overline{d} : a \succ \overline{a}$, \ 
$\overline{a} : \overline{b} \succ b$, 

\item $N_2$ results from $N_1$ by removing $\overline{b}$ and the 
preference statements
$a : b \succ \overline{b}$, \  $\overline{b} : \overline{c} \succ c$, 

\item $N_3$ results from $N_2$ by removing $\overline{c}$ and the 
preference statements $b : c \succ \overline{c}$ \ $\overline{c} : \overline{d} \succ d$, 

\item $N_4$ results from $N_3$ by removing $\overline{d}$ from the domain of $D$ and the 
preference statement $c : d \succ \overline{d}$.

\end{itemize}

Indeed, in each step the removed element is strictly dominated in the
considered CP-net.  So using the iterated elimination of strictly
dominated elements we reduced the original CP-net to one in which each
variable has a singleton domain and consequently found a unique
optimal outcome of the original CP-net $N$.

Finally, the following result shows that the introduced reduction relation
on CP-nets is complete for acyclic CP-nets.

\begin{theorem} \label{thm:acyclic}
For each acyclic CP-net $N$ a subnet $N'$ with the singleton domains exists such that
$N \myra^{*}_{\hspace{-1mm} NBR} N'$. The outcome associated with $N'$ is a unique
optimal outcome of $N$ and hence $N'$ is unique.
\end{theorem}
%\begin{changebar}
\Proof 
First note that if $N$ is an acyclic CP-net with some non-singleton domain, then
$N \myra_{\hspace{-1mm} NBR} N'$ for some subnet $N'$ of $N$.
Indeed, suppose $N$ is such a CP-net.
By acyclicity a variable $X$ exists with a non-singleton domain with
no parent variable that has a non-singleton domain.  So there exists in $N$
exactly one preference statement for $X$, say $X_I = a_I : \ \succ_i$,
where $X_I$ is the sequence of parent variables of $X$.  Reduce the
domain of $X$ to the maximal element in $\succ_i$.  Then for the
resulting subnet $N'$ we have $N \myra_{\hspace{-1mm} NBR} N'$.
Since $N'$ is also acyclic and has one variable less with a non-singleton domain, 
by iterating this procedure we obtain
a subnet $N'$ with the singleton domains such that
$N \myra^{*}_{\hspace{-1mm} NBR} N'$.

The claim that the outcome associated with
$N'$ is a unique optimal outcome of $N$ is a consequence of Theorem
\ref{thm:nets-ienbr}$(ii)$.
\HB

The singleton domains obtained via the use of the $\myra_{\hspace{-1mm} NBR}$ reduction 
correspond to the unique optimal outcome of an acyclic CP-net, as 
defined in \cite{BBHP.UAI99,BBHP.journal}.

\section{CP-net techniques in strategic games}
\label{sec:nets-to-games}

The established relationship between CP-nets and strategic games with
parametrized preferences also allows us to exploit the techniques
developed for the CP-nets when studying such games.

One natural idea is to consider a counterpart of the notion of an
acyclic CP-net.  We call a game with parametrized preferences
\oldbfe{hierarchical} if the CP-net $r({\cal N}(G))$ is acyclic.

We can introduce this notion directly, without using the CP-nets,
by considering a partition of players $1,\LL,n$ 
in the game
\[
(S_1, \LL, S_n, {\succ}(s_{-1}), \LL, {\succ}(s_{-n})),
\] 
where each $s_{-i}$ ranges over $S_{-i}$, into levels $1, \LL, k$
such that for each player $i$ at level $j$ and each $s_{-i} \in S_{-i}$
the preference ${\succ}(s_{-i})$ depends only on the entries in 
$s_{-i}$ associated with the players from levels $< j$.

So a game is hierarchical if the players can be partitioned into
levels $1,2,\ldots, k$, such that each player at level $j$ can express
his preferences without taking into account the players at his level or
higher levels (lower levels are more important).

We have then the following counterpart of Theorem \ref{thm:acyclic}.

\begin{theorem} \label{thm:acyclic1}
For each hierarchical game $G$ a subgame $G'$ with 
the singleton strategy sets exists such that
$G \myra^{*}_{\hspace{-1mm} NBR} G'$. The resulting joint strategy
associated with $G'$ is a unique Nash equilibrium 
of $G$ and hence $G'$ is unique.
\end{theorem}

\Proof
By an analogous argument as the one used in the proof of Theorem \ref{thm:acyclic}.
\HB
\VV

Given a hierarchical game $G$, by definition the CP-net $r({\cal
  N}(G))$ is acyclic. Thus we know that it has a unique optimal
outcome which can be found in linear time. This means that the unique
Nash equilibrium of $G$ can be found in linear time by the usual
CP-net techniques applied to $r({\cal N}(G))$.

Hierarchical games naturally represent multi-agent scenarios in which
agents (that is, players of the game) can be partitioned into levels
such that each agent can determine his preferences
without consulting agents at his level or lower levels.  Informally, agents at
one level  are `more important' than agents at lower levels  in the
sense that they can take their decisions without consulting them.

A more general class of games is obtained by analogy to graphical
games.  We define a \oldbfe{graphical game with parametrized
  preferences} as follows. Given a neighbour function \emph{neigh} we
assume that for each player $i$ and a joint strategy $s_i$ of his
opponents, the preference ${\succ}(s_{-i})$ depends only on the
entries in $s_{-i}$ associated with the players from
$\emph{neigh}(i)$.  Equivalently, we may just use the preference
relations ${\succ}_{s}^{i}$ for each player $i$ and each joint
strategy $s$ of the neighbours of $i$.  Hierarchical games are then
graphical games with parametrized preferences with acyclic neighbour
graphs.

Given a CP-net $N$ and the corresponding game ${\cal G}(N)$, the
dependency graph of $N$ uniquely determines the neighbour function
\emph{neigh} between the players in ${\cal G}(N)$. This allows us to
associate with each CP-net $N$ a graphical game with parametrized
preferences.  Conversely, each graphical game $G$ with parametrized
preferences uniquely determines a CP-net. It is obtained by proceeding
as in Section \ref{sec:to-nets} but by stipulating that the parent
relation corresponds to the neighbour function \emph{neigh}, that is,
by putting
\[
Pa(X_i) := \{X_j \mid j \in \emph{neigh}(i)\}.
\]

The counterparts of Theorems \ref{thm:N(G)} and \ref{thm:G(N)} then
hold for CP-nets and graphical games with parametrized preferences.

Note that we arrived at the concept of a hierarchical game through the
analogy with the acyclic CP-nets.  To see a natural example of such
games consider the problem of spreading a technology in a social
network, inspired by the problems studied in \cite{Mor00} for the case
of infinite number of players.  We assume that the players (users) are
connected in a network, which is a directed graph, and that there are
$k$ technologies (for example mobile telephone companies) $t_1, \LL, t_k$.
Assume further that each user, given two technologies,
prefers to use the one that is used by more of his neighbours in the network (for instance to cut
down on the telephone costs).

We model this situation as a graphical game with parametrized preferences.
We assume that each player $i$ has $k$ strategies, $t_1, \LL, t_k$,
and for each joint strategy $s$ of the neighbours of $i$
we define the preference relation ${\succ}_{s}^{i}$ by putting

\begin{equation}\label{equ:pref}
\begin{array}{l}
\mbox{$t_k {\succ}_{s}^{i} t_l$ iff $|s(t_k)| > |s(t_l)|$ or ($|s(t_k)| = |s(t_l)|$ and $k < l$),} 
\end{array}
\end{equation}
where $s(X)$ is the set of components of $s$ that are equal to the strategy $X$.
So we assume that in the case of a tie player $i$ prefers a technology with the lower index.

We can now analyze the process of selecting a technology by exploiting
the relation between hierarchical games and CP-nets.  Namely, suppose
that the above defined graphical game $G$ with parametrized
preferences is hierarchical. Then by virtue of Theorem
\ref{thm:acyclic1} $G \myra^{*}_{\hspace{-1mm} NBR} G'$, where in
$G'$ each player has a single strategy, $t_1$. The resulting joint
strategy is then a unique Nash equilibrium of $G$.  Additionally, by
the corresponding order independence result mentioned at the end of
Section \ref{sec:param}, $G'$ is a unique outcome of iterating the
$\myra_{\hspace{-1mm} NBR}$ reduction.

This corresponds to an informal statement that when the neighbour
function describes an acyclic graph, eventually technology $t_1$ is
adopted by everybody.  Because of the nature of the preference
relations used above, this result actually holds for a larger class of
graphical games with parametrized preferences.

\begin{figure}[htbp]
\begin{center} \ \setlength{\epsfxsize}{3.3cm}
\epsfbox{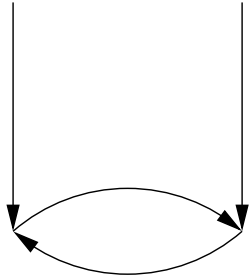}
\end{center}
\caption{A well-structured graph that is not acyclic}\label{fig:well}
\end{figure}

They correspond to the following class of directed graphs.  We call a
directed graph \oldbfe{well-structured} if levels can be assigned to
its nodes in such a way that each node has at least as many incoming
edges from the nodes with strictly lower levels than from the other
nodes. Of course, each directed acyclic graph is well-structured but
other examples exist, see, e.g., Figure \ref{fig:well}.

We have then the following result.

\begin{theorem} \label{thm:acyclic2}
  Consider a graphical game $G$ with parametrized preferences in which
  each player has $k$ strategies, $t_1, \LL, t_k$, the preference
  relations ${\succ}_{s}^{i}$ are defined by (\ref{equ:pref}), and the
  neighbour function describes a well-structured graph. Then $G
  \myra^{*}_{\hspace{-1mm} NBR} G'$, where in $G'$ each player has a
  single strategy, $t_1$, and the resulting joint strategy is a unique
  Nash equilibrium of $G$.
\end{theorem}
\Proof
We prove by induction on the level $m$ that
\begin{equation}
  \label{equ:level}
G \myra^{*}_{\hspace{-1mm} NBR} G',
\end{equation}
where in $G'$ each player of level $\leq m$
has a single strategy, $t_1$. This yields then the desired conclusion about Nash equilibrium by 
Theorem \ref{thm:ienbr}.

The claim holds for the lowest level, say $0$, as then each player of level $0$ has
no neighbours and hence his strategies $t_2, \LL, t_k$ can be eliminated as never best responses.

Suppose (\ref{equ:level}) holds for some level $m$. So we have $G \myra^{*}_{\hspace{-1mm} NBR} G'$, 
where in $G'$ each player of level $\leq m$ has a single strategy, $t_1$.
Consider the players of level $m+1$ in the game $G'$. Each of them has at least as many neigbours
with the single strategy $t_1$ than with other sets of strategies. So each joint strategy
of his neighbours has at least as many $t_1$s as other strategies. Hence 
$G' \myra^{*}_{\hspace{-1mm} NBR} G''$,
where in $G''$ each player of level $\leq m+1$ has a single strategy, $t_1$.
Consequently $G \myra^{*}_{\hspace{-1mm} NBR} G''$, which establishes the induction step.
\HB
\VV

The above example shows that graphical games with parametrized
preferences can be used to provide a natural qualitative analysis of
some problems studied in social networks.  Expressing the process of
selecting a technology using games with parametrized preferences, Nash
equilibria and elimination of never best responses is more natural
than using CP-nets. On the other hand we arrived at the relevant result about
adoption of a single technology by searching for an analogue of
Theorem \ref{thm:acyclic} about acyclic CP-nets.

\section{From SCSPs to graphical games}
\label{sec:soft-to}

In this and the next section we relate the notion of optimality 
in soft constraints and graphical games. 
To obtain an appropriate
match we assume that in graphical games payoffs are elements of a
linearly ordered set $A$ instead of the set of real numbers. (This
precludes the use of mixed strategies but they are not needed here.)
We denote then such games by 
\[
(S_1, \dots, S_n, neigh, p_1, \dots, p_n, A),
\]
where \emph{neigh} is the given neighbour function.

In this section we define two mappings from SCSPs to a specific
kind of graphical games.  In what follows we focus on SCSPs based on
c-semirings with the carrier linearly ordered by $\leq$ (e.g.~fuzzy or
weighted) and compare the concepts of optimal solutions in SCSPs with
Nash equilibria and Pareto efficient joint strategies in the graphical
games.  In both mappings we identify the players with the variables.
Since soft constraints link variables, the resulting game players are
naturally connected, which explains why we use graphical games.

\subsection{Local mapping}
\label{local}

Given a SCSP $P := \langle C, V, D, S \rangle$ we define the corresponding 
graphical game for $n=|V|$ players as follows:

\begin{itemize}
\item the players: one for each variable;
\item the strategies of player $i$: all values in the domain of the
  corresponding variable $x_i$;

\item the neighbourhood function: $j \in \emph{neigh}(i)$ iff the variables
$x_i$ and $x_j$ appear together in some constraint from $C$;

\item the payoff function of player $i$:
   
  Let $C_i \subseteq C$ be the set of constraints involving $x_i$ and
  let $X$ be the set of variables that appear together with $x_i$ in
  some constraint in $C_i$ (i.e., $X = \C{x_j \mid j \in \emph{neigh}(i)}$).
  Then given an assignment $s$ 
to all variables in $X \cup \C{x_i}$ the payoff
  of player $i$ w.r.t.~$s$ is defined by:
  
\[
p_i(s) :=\Pi_{c \in C_i} \text{def}_c(s \downarrow_{\text{con}_c}).
\]
\end{itemize}

We denote the resulting graphical game by $L(P)$ to emphasize 
the fact that the payoffs are
obtained using \emph{local} information about each variable, by looking
only at the constraints in which it is involved.   

One could think of a different mapping where players correspond 
to constraints. However, such a mapping can be obtained by applying the 
local mapping $L$ to the hidden variable encoding \cite{hidden} of the SCSP in input. 

\subsubsection{General case}

In general, the concepts of optimal solutions of a SCSP
$P$ and the Nash equilibria of the 
derived game $L(P)$ are unrelated.
Indeed, consider the fuzzy CSP defined 
at the end of Section \ref{subsec:soft}.
The corresponding game has:

\begin{itemize}
\item three players, $x$, $y$, and $z$;
\item each player has two strategies, $a$ and $b$;
\item the neighbourhood function is defined by:
\[
\emph{neigh}(x):=\{y \}, \ \emph{neigh}(y):=\{x,z \}, \ \emph{neigh}(z):=\{ y\};
\]
\item the payoffs of the players are defined as follows:
\begin{itemize}
\item for player $x$: 

$p_x(aa*):=0.4$, $p_x(ab*):=0.1$, $p_x(ba*):=0.3$, $p_x(bb*):=0.5$;

\item for player $y$: 

$p_y(aaa):=0.4$, $p_y(aab):=0.3$, $p_y(abb):=0.1$, $p_y(bbb):=0.5$, 

$p_y(bba):=0.5$, $p_y(baa):=0.3$, $p_y(bab):=0.3$, $p_y(aba):=0.1$;
     
\item for player $z$: 

$p_z(*aa):=0.4$, $p_z(*ab):=0.3$, $p_z(*ba):=0.1$, $p_z(*bb):=0.5$;
\end{itemize}
\end{itemize}
where $*$ stands for either $a$ or $b$ and where to facilitate the
analysis we use the canonical extensions of the payoff functions
$p_x$ and $p_z$ to the functions on $\C{a,b}^3$.

This game has two Nash equilibria: $aaa$ and $bbb$. However, only
$bbb$ is an optimal solution of the fuzzy SCSP.

One could thus think that in general
the set of Nash equilibria is a superset of the
set of optimal solutions of the corresponding 
SCSP. However, this is not the case. Indeed, 
consider a fuzzy CSP with as before three variables, $x, y$ and $z$, 
each with the domain $\C{a,b}$, but now with the constraints:
\II

$C_{xy} := \{(aa, 0.9), (ab, 0.6), (ba, 0.6), (bb, 0.9)\}$,

$C_{yz} := \{(aa, 0.1), (ab, 0.2), (ba, 0.1), (bb, 0.2)\}$.
\II

Then $aab, \ abb, \ bab$ and $bbb$ are all optimal solutions but only
$aab$ and $bbb$ are Nash equilibria of the corresponding graphical game.

\subsubsection{SCSPs with strictly monotonic combination} 
\label{subsub:mon}

Next, we consider the case when the multiplicative operator $\times$ is
strictly monotonic.  Recall that given a c-semiring $\langle A,+,
\times, \0, \1 \rangle$, the operator $\times$ is \oldbfe{strictly
  monotonic} if for any $a,b,c \in A$ such that $a<b$ we have $c
\times a < c \times b$.  (The symmetric condition is taken care of by
the commutativity of $\times$.)  

Note for example that in weighted CSP $\times$ is strictly monotonic,
as $a < b$ in the carrier means that $b < a$ as reals, so for any $c$
we have $c + b < c + a$, i.e., $c \times a < c \times b$ in the carrier.
In contrast, the 
fuzzy CSPs $\times$ are not strictly
monotonic, as $a < b$ does not imply that $min(a,c) < min(b,c)$ for
all $c$.  

So consider now a c-semiring with a linearly ordered carrier and a 
strictly monotonic multiplicative operator. As in the previous case,
given an SCSP $P$, it is possible that a Nash equilibrium of $L(P)$ is
not an optimal solution of $P$. Consider for example a
weighted SCSP $P$ with
\begin{itemize}
\item two variables, $x$ and $y$, each with the
domain $D=\{a,b\}$;
\item one constraint $C_{xy} := \{(aa, 3), (ab, 10), (ba, 10), (bb, 1)\}$.
\end{itemize}
The corresponding game $L(P)$ has:

\begin{itemize}
\item two players, $x$ and $y$, who are neighbours of each other;

\item each player has two strategies, $a$ and $b$;

\item the payoffs defined by: 

$p_x(aa):=p_y(aa):=7$, $p_x(ab):=p_y(ab):=0$, 

$p_x(ba):=p_y(ba):=0$, $p_x(bb):=p_y(bb):=9$.
\end{itemize}

Notice that, in a weighted CSP we have $a \leq b$ in the carrier iff
$b \leq a$ as reals, so when passing from the SCSP to the
corresponding game, we have complemented the costs w.r.t.~10, when
making them payoffs. In general, given a weighted CSP, we can define
the payoffs (which must be maximized) from the costs (which must be
minimized) by complementing the costs w.r.t.~the greatest cost used in
any constraint of the problem.

Here $L(P)$ has two Nash equilibria, $aa$ and $bb$, but only $bb$ is
an optimal solution.  Thus, as in the fuzzy case, we have that there
can be a Nash equilibrium of $L(P)$ that is not an optimal solution of
$P$.  However, in contrast to the fuzzy case, 
the set of Nash equilibria
of $L(P)$ is now a superset of the set of optimal solutions of $P$.
In fact, a stronger result holds.

\begin{theorem}
Consider a SCSP $P$ defined on a c-semiring $\langle A,+,\times,\0,\1
\rangle$, where $A$ is linearly ordered and $\times$ is 
strictly monotonic, and the corresponding graphical game $L(P)$. 
Then 

\begin{enumerate} \smallromani
\item Every optimal solution of $P$ is a Nash equilibrium of $L(P)$.  

\item Every optimal solution of $P$ is a Pareto efficient joint strategy
in $L(P)$.
\end{enumerate}

\end{theorem}

\Proof 

\NI
$(i)$ We prove that if a joint strategy $s$ 
is not a Nash equilibrium of game $L(P)$, then it is not an optimal 
solution of SCSP $P$. 

Let $a$ be the strategy of player $x$ in $s$, and let
$s_{\emph{neigh}(x)}$ and $s_{Y}$ 
be, respectively, the joint strategy of the neighbours of $x$, and of
all other players, in $s$. That is, $V = \{x\} 
\cup \emph{neigh}(x) \cup Y$ and we write $s$ as
$(a, s_{\emph{neigh}(x)}, s_Y)$.

By assumption there is a strategy $b$ for $x$ such that the payoff 
$p_{x}(s')$ for the joint strategy 
$s' := (b, s_{\emph{neigh}(x)}, s_Y)$ is higher than 
$p_x(s)$. (We use here the canonical extension of $p_x$ to the Cartesian product of
all the strategy sets).

So by the definition of the mapping $L$
\[
\Pi_{c \in C_x} \text{def}_c(s \downarrow_{\text{con}_c})< \Pi_{c \in C_x}
\text{def}_c(s' \downarrow_{\text{con}_c}),
\]
where $C_x$ is the set of all the constraints involving $x$ in SCSP $P$. 
But the preference of $s$ and $s'$ is the same on all the
constraints not involving $x$ and $\times$ is strictly monotonic,
so we conclude that
\[
\Pi_{c \in C} \text{def}_c(s \downarrow_{\text{con}_c})< \Pi_{c \in C}
\text{def}_c(s' \downarrow_{\text{con}_c}).
\]
This means that $s$ is not an optimal solution of $P$.
\II

\NI
$(ii)$
We prove that if a joint strategy $s$ is not Pareto efficient in the
game $L(P)$, then it is not an optimal 
solution of SCSP $P$. 

Since $s$ is not Pareto efficient, there is a joint strategy 
$s'$ such that $p_i(s) \leq p_i(s')$ for all $i \in [1..n]$ and 
$p_i(s) < p_i(s')$ for some $i \in [1..n]$.
Let us denote with $I = \{i \in [1..n]$ such that $p_i(s) < p_i(s')\}$.
By the definition of the mapping $L$, we have:
\[
\Pi_{c \in C_i} \text{def}_c(s \downarrow_{\text{con}_c})< \Pi_{c \in C_i}
\text{def}_c(s' \downarrow_{\text{con}_c}),
\]
for all $i \in I$ and where $C_i$ is the set of all 
the constraints involving the variable corresponding to player $i$ in SCSP $P$. 
Since the preference of $s$ and $s'$ is the same on all the
constraints not involving any $i \in I$, and since $\times$ is strictly monotonic,
we have:
\[
\Pi_{c \in C} \text{def}_c(s \downarrow_{\text{con}_c})< \Pi_{c \in C}
\text{def}_c(s' \downarrow_{\text{con}_c}).
\]
This means that $s$ is not an optimal solution of $P$.
\HB \VV

To see that there may be joint strategies that are both Nash equilibria and Pareto efficient
but do not correspond to the optimal solutions, consider a
weighted SCSP $P$ with
\begin{itemize}
\item two variables, $x$ and $y$, each with 
domain $D=\{a,b\}$;
\item constraint $C_{x} := \{(a, 2), (b, 1)\}$;
\item constraint $C_{y} := \{(a, 4), (b, 7)\}$;
\item constraint $C_{xy} :=\{(aa,0), (ab, 10), (ba, 10), (bb,0) \}$.
\end{itemize}
The corresponding game $L(P)$ has:

\begin{itemize}
\item two players, $x$ and $y$, who are neighbours of each other;

\item each player has two strategies: $a$ and $b$;

\item the payoffs defined by: 
$p_x(aa):=8$, 
$p_y(aa):=6$, 
$p_x(ab):=p_y(ab):=0$, 
$p_x(ba):=p_y(ba):=0$, 
$p_x(bb):= 9$,
$p_y(bb):=3$.
\end{itemize}

As above, when passing from an SCSP to the
corresponding game, we have complemented the costs w.r.t.~10, when
turning them to payoffs. 
%In general, given a weighted CSP, we can define
%the payoffs (which must be maximized) from the costs (which must be
%minimized) by complementing the costs w.r.t.~the greatest cost used in
%any constraint of the problem.
$L(P)$ has two  Nash equilibria: $aa$ and $bb$. 
They are also both Pareto efficient.
However, only $aa$ is optimal in $P$.

\subsubsection{Classical CSPs}

Note that in the classical CSPs $\times$ is not strictly monotonic, as $a < b$ implies
that $a = 0$ and $b = 1$ but $c \land a < c \land b$ does not
hold then for $c = 0$.  In fact, 
the above result does not hold for classical CSPs.
Indeed, consider a CSP with:
\begin{itemize}
\item three variables: $x$, $y$, and $z$, each with the domain $\{a,b\}$;
\item two constraints: $C_{xy}$ (over $x$ and $y$) and $C_{yz}$ (over
  $y$ and $z$) defined by:

$C_{xy} := \{(aa, 1), (ab, 0), (ba, 0), (bb, 0)\}$,

$C_{yz} := \{(aa, 0), (ab, 0), (ba, 1), (bb, 0)\}$.
\end{itemize}

This CSP has no solutions in the classical sense, i.e., each optimal solution, in particular $baa$, 
has preference 0. However,
$baa$ is not a Nash equilibrium of the resulting graphical game, since
the payoff of player $x$ increases when he switches to the strategy $a$.

On the other hand, if we
restrict the domain of $L$ to consistent CSPs, that is, CSPs with at least one
solution with value 1, then it yields games in which the set
of Nash equilibria that are also Pareto efficient joint strategies
coincides with the set of solutions of the CSP.

\begin{theorem} \label{thm:CSP}
Consider a consistent CSP $P$ and the corresponding graphical game $L(P)$. 
Then an instantation of the variables of $P$ is a 
solution of $P$ iff it is a Nash equilibrium and Pareto efficient joint strategy
in $L(P)$.
\end{theorem}
\Proof 
Consider a solution $s$ of $P$. In the resulting game $L(P)$ the payoff
to each player is maximal, namely 1. So the 
joint strategy $s$ is both a Nash equilibrium and Pareto efficient.
Conversely, every Pareto efficient joint strategy in $L(P)$ corresponds to solution of $P$.
\HB 
\VV

There are other ways to relate CSPs and games so that the CSP solutions
and the Nash equilibria coincide. This is what is done in
\cite{gottlob}, where a mapping from the strategic games to CSPs is defined.
Notice that our mapping goes in the opposite direction and it is not
the reverse of the one in \cite{gottlob}. In fact, the mapping in
\cite{gottlob} is not reversible.

\subsection{Global mapping}
\label{global}

The mapping $L$ is in some sense `local', since it considers the
neighbourhood of each variable. An alternative `global' mapping
considers all constraints.  More precisely, given a SCSP $P=\langle C,
V, D, S \rangle$, with a linearly ordered carrier $A$ of $S$, we
define the corresponding game on $n=|V|$ players, $GL(P)=(S_1, \dots,
S_n, p_1, \dots, p_n, A)$ by using the following payoff function $p_i$
for player $i$:

\begin{itemize}

\item 
given an assignment $s$ to \emph{all} variables in $V$
\[
p_i(s) := \Pi_{c \in C} \text{def}_c(s \downarrow_{\text{con}_c}).
\]
\end{itemize}

Notice that in the resulting game the payoff functions of all 
players are the same.
Then the following result analogous to Theorem \ref{thm:CSP} holds.

\begin{theorem}
\label{10}
Consider an SCSP $P$ over a linearly ordered carrier,
and the corresponding graphical game $GL(P)$. 
Then an instantiation of the variables of $P$ is an optimal 
solution of $P$ iff it is a Nash equilibrium and Pareto efficient in $GL(P)$.

\end{theorem}

\Proof
An optimal solution of $P$, say $s$, 
is a joint strategy for which all players have the same, highest, payoff.
So no other joint strategy exists for which some 
player is better off and consequently $s$ is both a Nash equilibrium and Pareto efficient.
Conversely, every Pareto efficient joint strategy in $GL(P)$ has the highest payoff, so it
corresponds to an optimal solution of $P$.
\HB 
\VV

The global mapping $GL$ has the advantage of providing a precise
relationship between the optimal solutions and joint strategies
that are both Nash equilibria and Pareto efficient. However, it has an
obvious disadvantage from the computational point of view, since it
requires to consider all the complete assignments of the SCSP.

% \subsection{Summary of results}

% Summarizing, in this section we have analyzed the relationship between 
% the optimal solutions of SCSPs and the Nash equilibria of graphical games. 
% In \cite{gottlob} CSPs have been shown to be sufficient to model Nash equilibria of graphical games. 
% Here we have considered the question whether the Nash equilibria of graphical games can model the 
% optimal solutions of SCSPs. We have provided two mappings from SCSPs to graphical games,
% showing that (with some conditions for the local mapping) 
% the set of Nash equilibria of the obtained game contains the optimal solutions of the given SCSP.

% Nash equilibria can be seen as the optimal elements in very specific
% orderings, where dominance is based on exactly one change in the joint
% strategy, while SCSPs can model any ordering. So we conjecture that it
% is not possible to find a mapping from SCSPs to the graphical games
% for which the optimals coincide with Nash equilibria.  Such a
% conjecture is also supported by Theorem \ref{thm:G(N)} of Section
% \ref{sec:to-games} that states that the optimals of a CP-net coincide
% with the Nash equilibria, whereas the CP-nets can model strictly less
% orderings than the SCSPs, see \cite{prest}.
 
\section{From graphical games to SCSPs}
\label{sec:to-soft}

Next, we define a mapping from graphical games to SCSPs.
To define it
we limit ourselves to SCSPs defined on c-semirings which 
are the Cartesian product of linearly ordered c-semirings (see Section \ref{subsec:soft}).

\subsection{The mapping}

Given a graphical game $G=(S_1, \dots, S_n, \emph{neigh}, p_1, \dots, p_n, A)$
we define the corresponding SCSP $L'(G)= \langle C, V, D, S \rangle$, 
as follows:
\begin{itemize}

\item each variable $x_i$ corresponds to a player $i$;

\item the domain $D(x_i)$ of the variable $x_i$ consists of
  the set of strategies of player $i$, i.e., $D(x_i) : = S_i$;

\item the c-semiring is 

$\langle A_1 \times \cdots \times A_n, (+_1, \dots, +_n),
(\times_1, \dots, \times_n),(\0_1, \dots, \0_n),(\1_1, \dots, \1_n) \rangle$,

\NI
the Cartesian product of $n$ \emph{arbitrary} linearly ordered semirings;

\item soft constraints: for each variable $x_i$, one constraint $\langle
  \text{def},\text{con} \rangle$ such that:
\begin{itemize}
\item $\text{con} = \emph{neigh}(x_i) \cup \{x_i \}$;
\item $\text{def}: \Pi_{y \in \text{con}} D(y) \rightarrow  A_1 \times \cdots
  \times A_n$ such that for any $s \in \Pi_{y \in \text{con}} D(y)$, 
  $\text{def}(s) :=(d_1, \dots, d_n)$ with $d_j=\1_j$ for every $j \neq i$ and 
  $d_i=f(p_i(s))$, where 
$f: A \rightarrow A_i$ is an order preserving
mapping from payoffs to preferences
(i.e., if $r>r'$ then $f(r)>f(r')$ in the c-semiring's ordering). 
\end{itemize}
\end{itemize}  

To illustrate it consider again the previously used Prisoner's Dilemma game:

\begin{center}
\begin{game}{2}{2}
       & $C_2$    & $N_2$\\
$C_1$   &$3,3$   &$0,4$\\
$N_1$   &$4,0$   &$1,1$
\end{game}
\end{center}

Recall that in this game the only Nash equilibrium is 
$(N_1,N_2)$, while the other three joint strategies are 
Pareto efficient.

We shall now construct a corresponding SCSP based on 
the Cartesian product of two weighted semirings.
This SCSP according to the mapping $L'$ has:\footnote{Recall that in the weighted semiring
\textbf{1} equals 0.}
\begin{itemize}
\item two variables: $x_1$ and $x_2$, each with the domain
$\{c,n\}$;
\item two constraints, both on $x_1$ and $x_2$:
\begin{itemize}
\item constraint $c_1$ with 
$\text{def}(cc) := \langle 7, 0 \rangle$, 
$\text{def}(cn) := \langle 10, 0 \rangle$, 
$\text{def}(nc) := \langle 6, 0 \rangle$, 
$\text{def}(nn) := \langle 9, 0 \rangle$;
\item constraint $c_2$ with 
$\text{def}(cc) := \langle 0, 7 \rangle$, 
$\text{def}(cn) := \langle 0, 6 \rangle$, 
$\text{def}(nc) := \langle 0, 10 \rangle$, 
$\text{def}(nn) := \langle 0, 9 \rangle$;
\end{itemize}
\end{itemize}

The optimal solutions of this SCSPs are: $cc$, with preference
$\langle 7,7 \rangle$, $nc$, with preference $\langle 10,6 \rangle$,
$cn$, with preference $\langle 6,10 \rangle$.  The remaining solution,
$nn$, has a lower preference in the Pareto ordering. Indeed, its
preference $\langle 9,9 \rangle$ is dominated by $\langle 7,7 \rangle$, 
the preference of $cc$ (since
preferences are here costs and have to be minimized).  Thus the
optimal solutions coincide here with the Pareto efficient joint strategies
of the given game. This is true in general.

\begin{theorem}
\label{t1}
Consider a graphical game $G$ and a corresponding SCSP $L'(G)$.
Then the optimal solutions of $L'(G)$ coincide with 
the Pareto efficient joint strategies of $G$.
\end{theorem}

\Proof
%Let us denote with $Pe(G)$ the set of Pareto efficient strategies of 
%$G$ and with $Opt(L'(G))$.
%$Pe(G) \subseteq Opt(L'(G))$. In fact, 
%given $s \in Pe(G)$, if some player $i$ there exists a strategy $s'$ such 
%that $p_i(s')>p_i(s)$ then there also exists another player $j$ such that 
%$p_j(s')<p_j(s)$. 
In the definition of the mapping $L'$ we stipulated that the mapping
$f$ maintains the ordering from the payoffs to preferences.
As a result
each joint strategy $s$ corresponds to the $n$-tuple 
of preferences $(f(p_1(s)), \dots, f(p_n(s)))$
and the Pareto orderings on the $n$-tuples  $(p_1(s), \dots, p_n(s))$
and $(f(p_1(s)), \dots,$ $f(p_n(s)))$ coincide.
Consequently a sequence $s$
is an optimal solution of the SCSP $L'(G)$ iff
$(f(p_1(s)), \dots, f(p_n(s)))$ is a maximal element 
of the corresponding Pareto ordering. 
\HB \VV

We notice that $L'$ is injective and, thus, can be reversed on its image.
When such a reverse mapping is applied to these specific SCSPs,
payoffs correspond to projecting of the players' valuations to a subcomponent.

%From the computational point of view, this means that, to find 
%the Pareto efficient strategies of a game, it is enough to 
%map such a game into an SCSP and then to solve such an SCSP.
%If the given game is graphical and with neighbourhoods with bounded size, 
%then the mapping is polynomial. Thus the cost of the mapping is 
%absorbed by the solution phase. Tractable classes of SCSPs  
%can help in making this second phase polynomial as well.

\subsection{Pareto efficient Nash equilibria}

As mentioned earlier, in \cite{gottlob} a mapping is defined
from the graphical
games to CSPs such that Nash equilibria coincide with the solutions of CSP.  
Instead, our mapping is from the graphical games to
SCSPs, and is such that Pareto efficient joint strategies
and the optimal solutions coincide.

Since CSPs can be seen as a special instance of SCSPs, where only
\textbf{1}, \textbf{0}, the top and bottom elements of the semiring,
are used, it is possible to add to any SCSP a set of hard constraints.
Therefore we can merge the results of the two mappings into a single 
SCSP, which contains the soft constraints generated by $L'$ 
and also the hard constraints generated by the mapping in \cite{gottlob},
Below we denote these hard constraints by $H(G)$.
%brent r2
We recall that 
each constraint in $H(G)$ corresponds to a player, 
has the variables corresponding 
to the player and it neighbours and allows only tuples corresponding to the
strategies in which the player has no so-called regrets.
If we do this, then the optimal solutions of the new SCSP
with preference higher than \textbf{0}
are the Pareto efficient Nash equilibria of the given game, that is, 
those Nash equilibria which dominate or are 
incomparable with all other Nash equilibria according  
to the Pareto ordering. Formally, we have the following result.
  
\begin{theorem}
Consider a graphical game $G$ and the SCSP $L'(G) \cup H(G)$.
If the optimal solutions of $L'(G) \cup H(G)$
have global preference greater than \textbf{0},
they correspond to the 
Pareto efficient Nash equilibria of $G$.
\end{theorem}

\Proof 
Given any solution $s$, let $p$ be its preference in $L'(G)$
and $p'$ in $L'(G) \cup H(G)$. By the construction of the constraints
$H(G)$ we have that $p'$ equals $p$ if $s$ is a Nash equilibrium and
$p'$ equals \textbf{0} otherwise.  The remainder of the argument is as
in the proof of Theorem \ref{t1}.  \HB \VV

For example, in the Prisoner's Dilemma game,
the mapping in \cite{gottlob} would generate just one constraint
on $x_1$ and $x_2$ with $nn$ as the only allowed tuple.
In our setting, when using as the linearly ordered c-semirings
the weighted semirings,
this would become a soft constraint
with 
\[
\text{def}(cc) := \text{def}(cn) := \text{def}(nc) = \langle \infty, \infty \rangle, \
\text{def}(nn) := \langle 0, 0 \rangle.
\]
With this new constraint, all solutions have the preference 
$\langle \infty, \infty \rangle$, except for $nn$ which has the
preference $\langle 9,9 \rangle$ and thus is optimal.  
This solution corresponds to the joint strategy $(N_1,N_2)$ with the payoff $(1,1)$ 
(and thus preference $(9,9)$).
%brent r2
This is the only Nash equilibrium and thus the only Pareto efficient Nash equilibrium. 

%The computational aspects of this mapping follows 
%a similar reasoning as above: finding 
%the Pareto efficient Nash equilibria is as hard 
%as solving and SCSP, if the neighbourhood size is bounded.
 
This method allows us to identify among Nash equilibria the `optimal' ones.
One may also be interested in knowing whether there exist
Nash equilibria which are also Pareto efficient joint strategies.
For example, in the Prisoners' Dilemma example, there are no such Nash equilibria.
To find any such joint strategies we can use the two mappings separately, 
to obtain, given a game $G$, both an SCSP $L'(G)$ and a CSP $H(G)$
(using the mapping in \cite{gottlob}). Then we should take the intersection 
of the set of optimal solutions of $L'(G)$ and the set of solutions of 
$H(G)$. 

% \subsection{Summary of results}

% We have considered the relationship between optimal solutions of SCSPs
% and Pareto efficient joint strategies in graphical games.  The local mapping of Section
% \ref{local} turns out to map optimal solutions of a given SCSP to
% Pareto efficient joint strategies, while the global mapping of Section \ref{global}
% yields a one-to-one correspondence.  For the reverse direction it is
% possible to define a mapping such that these two notions of optimality
% coincide. However, none of these mappings are onto.

\section{Conclusions}
\label{conc}

In this paper we related three formalisms that are commonly used to
reason about optimal outcomes: strategic games, CP-nets and soft
constraints. To this end we modified the concept of strategic games
to games with parametrized preferences and showed that the optimal
outcomes in CP-nets are exactly Nash equilibria of such games.  This
allowed us to exploit game-theoretic techniques in search for the optimal
outcomes of CP-nets.
In the other direction, we showed how the notion of an acyclic CP-net
naturally leads to the concept of a hierarchical game. Such games
have a unique Nash equilibrium.

We also considered the relation between graphical games and various
classes of soft constraints. While for soft constraints there is only
one notion of optimality, for graphical games there are at least two.
In this paper we have considered Nash equilibria and Pareto efficient
joint strategies.  We showed that for a natural (local) mapping from
soft CSPs to graphical games in general no relation exists between the
notions of optimal solutions of soft CSPs and Nash equilibria.  On the
other hand, when in the SCSPs the preferences are combined using a
strictly monotonic operator, the optimal solutions of the SCSP are
included both in the Nash equilibria of the game and in the set of
Pareto efficient joint strategies.  In general the inclusions cannot
be reversed.  We have also exhibited a (global) mapping from the
graphical games to a class of SCSPs such that the Pareto efficient
joint strategies of the game coincide with the optimal solutions of
the SCSP.

For the reverse direction we showed that for a natural mapping from
the graphical games to a class of SCSPs the optimal solutions coincide
with Pareto efficient joint strategies.  Moreover, if we add suitable
hard constraints to the soft constraints, optimal solutions coincide
with Pareto efficient Nash equilibria.

The results of this paper clarify the relationship between various
notions of optimality used in strategic games, CP-nets and soft
constraints.  These results can be used in a number of ways. One obvious way
is to try to exploit computational results existing for one of these
areas in another.  This has been pursued already in \cite{gottlob} for
games versus hard constraints. Using our results this can also be done
for strategic games versus CP nets or soft constraints.  For example,
finding a Pareto efficient joint strategy involves mapping a game into
a soft CSP and then solving it.  Similar
approach can also be applied to Pareto efficient Nash equilibria,
which can be found by solving a suitable soft CSP.

\section*{Acknowledgements}

We thank the reviewers for helpful comments and suggestions.

%\bibliographystyle{plain}
%\bibliographystyle{handbk}

%\bibliography{/ufs/apt/bib/rossi05,/ufs/apt/bib/clp2,/ufs/apt/bib/e,/ufs/apt/bib/e1}
%\bibliography{/ufs/apt/bib/rossi05,/ufs/apt/bib/clp2,/ufs/apt/bib/e1}

%\bibliography{/ufs/apt/e1}
%\bibliography{rossi05,/home/staff/apt/bib/e,/home/staff/apt/bib/bibliophd-short}

\begin{thebibliography}{10}

\bibitem{Apt05}
K.~R. Apt.
\newblock Rationalizability and order independence.
\newblock In {\em Proc. 10th Conference on Theoretical Aspects of Reasoning
  about Knowledge (TARK '05)}, pages 22--38.
\newblock Available from \url{http://portal.acm.org}.


\bibitem{ARV05}
K.~R. Apt, F.~Rossi, and K.~B. Venable.
\newblock CP-nets and Nash equilibria.
\newblock In {\em Proc. of the Third International Conference on Computational
  Intelligence, Robotics and Autonomous Systems (CIRAS '05)}, pages 1--6.
\newblock Available from \url{http://arxiv.org/abs/cs/0509071}.

\bibitem{ARV08}
K.~R. Apt, F.~Rossi, and K.~B. Venable.
\newblock A comparison of the notions of optimality in 
soft constraints and graphical games.
\newblock In {\em Recent Advances in Constraints}, 
\newblock In Lecture Notes in Computer Science, Springer, 2008.
\newblock To appear. 
%Available from \url{http://arxiv.org/abs/cs/0509071}.


\bibitem{jacm}
S.~Bistarelli, U.~Montanari, and F.~Rossi.
\newblock Semiring-based constraint solving and optimization.
\newblock {\em Journal of the ACM}, 44(2):201--236, mar 1997.

\bibitem{BBHP.journal}
C.~Boutilier, R.~I. Brafman, C.~Domshlak, H.~H. Hoos, and D.~Poole.
\newblock {CP}-nets: A tool for representing and reasoning with conditional
  ceteris paribus preference statements.
\newblock {\em J. Artif. Intell. Res. (JAIR)}, 21:135--191, 2004.

\bibitem{BBHP.UAI99}
C.~Boutilier, R.~I. Brafman, H.~H. Hoos, and D.~Poole.
\newblock Reasoning with conditional ceteris paribus preference statements.
\newblock In K.~B. Laskey and H.~Prade, editors, {\em UAI '99: Proceedings of
  the Fifteenth Conference on Uncertainty in Artificial Intelligence,
  Stockholm, Sweden, July 30-August 1}, pages 71--80. Morgan Kaufmann, 1999.

\bibitem{dimo}
R.~Brafman and Y.~Dimopoulos.
\newblock Extended semantics and optimization algorithms for {CP}-networks.
\newblock {\em Computational Intelligence}, 20(2):218-- 245, 2004.

\bibitem{fuzzy1}
H.~Fargier, D.~Dubois, and H.~Prade.
\newblock The calculus of fuzzy restrictions as a basis for flexible constraint
  satisfaction.
\newblock In {\em IEEE International Conference on Fuzzy Systems}, 1993.

\bibitem{Domshlak:Brafman:kr02}
C.~Domshlak and R.~I. Brafman.
\newblock {CP}-nets: Reasoning and consistency testing.
\newblock In D.~Fensel, F.~Giunchiglia, D.~L. McGuinness, and M.~Williams,
  editors, {\em Proceedings of the Eight International Conference on Principles
  and Knowledge Representation and Reasoning (KR-02), Toulouse, France, April
  22-25}, pages 121--132. Morgan Kaufmann, 2002.

\bibitem{gottlob}
G.~Greco, G.~Gottlob, and F.~Scarcello.
\newblock Pure {Nash} equilibria: Hard and easy games.
\newblock {\em J. of Artificial Intelligence Research}, 24:357--406, 2005.

\bibitem{GKZ90}
I.~Gilboa, E.~Kalai, and E.~Zemel.
\newblock On the order of eliminating dominated strategies.
\newblock {\em Operation Research Letters}, 9:85--89, 1990.


\bibitem{constNash}
G.~Greco and F.~Scarcello.
\newblock Constrained Pure Nash Equilibria in Graphical Games,
\newblock Proceedings of the 16th Eureopean Conference on Artificial
               Intelligence (ECAI'2004), pages 181--185, IOS Press, 2004.  

\bibitem{KLS01}
M.~Kearns, M.~Littman, and S.~Singh.
\newblock Graphical models for game theory.
\newblock In {\em Proceedings of the 17th Conference in Uncertainty in
  Artificial Intelligence (UAI '01)}, pages 253--260. Morgan Kaufmann, 2001.

\bibitem{tambe}
R. T.~Maheswaran, J. P.~Pearce, and M.~Tambe. 
\newblock Distributed Algorithms for DCOP: A Graphical-Game-Based Approach. 
\newblock Proceedings of the ISCA 17th International Conference on
Parallel and Distributed Computing Systems (ISCA PDCS 2004), pages 432--439, ISCA, 2004.


\bibitem{hidden}
N. Mamoulis and K. Stergiou.
\newblock Solving non-binary CSPs using the hidden variable encoding.
\newblock In Lecture Notes in Computer Science volume 2239, Springer, 2001.

\bibitem{Mor00}
S.~Morris.
\newblock Contagion.
\newblock {\em The Review of Economic Studies}, 67(1):57--78, 2000.

\bibitem{Mye91}
R.~B. Myerson.
\newblock {\em Game Theory: Analysis of Conflict}.
\newblock Harvard Univ Press, Cambridge, Massachusetts, 1991.

\bibitem{OR94}
M.~J. Osborne and A.~Rubinstein.
\newblock {\em A Course in Game Theory}.
\newblock The {MIT} Press, Cambridge, Massachusetts, 1994.

\bibitem{MRS06}
F.~Rossi, P.~Meseguer and T.~Schiex.
\newblock Soft constraints.
\newblock In T.~Walsh F.~Rossi, P. Van~Beek, editor, {\em Handbook of
  Constraint programming}, pages 281--328. Elsevier, 2006.

\bibitem{aaai05}
S.D. Prestwich, F.Rossi, K.B. Venable, and T.~Walsh.
\newblock Constraint-based preferential optimization.
\newblock In {\em AAAI}, pages 461--466, 2005.

\bibitem{ruttkay-fuzzy}
Z.~Ruttkay.
\newblock Fuzzy constraint satisfaction.
\newblock In {\em Proceedings 1st {IEEE} Conference on Evolutionary Computing},
  pages 542--547, Orlando, 1994.

\bibitem{Ste90}
M.~Stegeman.
\newblock Deleting strictly eliminating dominated strategies.
\newblock Working Paper 1990/6, Department of Economics, University of North
  Carolina, 1990.

\end{thebibliography}

\end{document}